\documentclass[runningheads]{llncs}

\usepackage{eccv}

\usepackage{eccvabbrv}

\usepackage{graphicx}
\usepackage{booktabs}

\usepackage{bm}
\usepackage{colortbl}
\usepackage{xcolor}

\usepackage[accsupp]{axessibility}  

\usepackage[pagebackref,breaklinks,colorlinks,citecolor=eccvblue]{hyperref}

\usepackage{orcidlink}

\definecolor{best}{rgb}{0.0,0.5,0.5} 
\definecolor{todo_color}{rgb}{.7,.1,.1}

\newcommand{\vect}[1]{\boldsymbol{\mathbf{#1}}}

\newcommand{\losslam}[1]{\lambda_{\text{#1}}}
\newcommand{\lossL}[1]{\mathcal{L}_{\text{#1}}}
\newcommand{\lossterm}[1]{\losslam{#1}\lossL{#1}}

\begin{document}

\title{Autoregressive Appearance Prediction for 3D Gaussian Avatars} 



\author{Michael Steiner\inst{1,2}\thanks{Work done during an internship at Meta} \and
Zhang Chen\inst{2} \and
Alexander Richard\inst{2} \and
Vasu Agrawal\inst{2} \and
Markus Steinberger\inst{1} \and
Michael Zollhoefer\inst{2}
}

\authorrunning{M.~Steiner et al.}

\institute{Graz University of Technology, Austria \and
Meta Reality Labs, Pittsburgh, USA \\
\url{https://steimich96.github.io/AAP-3DGA/}}

\maketitle

\begin{abstract}
    A photorealistic and immersive human avatar experience demands capturing fine, person-specific details such as cloth and hair dynamics, subtle facial expressions, and characteristic motion patterns.
    Achieving this requires large, high-quality datasets, which often introduce ambiguities and spurious correlations when very similar poses correspond to different appearances.
    Models that fit these details during training can overfit and produce unstable, abrupt appearance changes for novel poses.
    We propose a 3D Gaussian Splatting avatar model with a spatial MLP backbone that is conditioned on both pose and an appearance latent.
    The latent is learned during training by an encoder, yielding a compact representation that improves reconstruction quality and helps disambiguate pose-driven renderings.
    At driving time, our predictor autoregressively infers the latent, producing temporally smooth appearance evolution and improved stability.
    Overall, our method delivers a robust and practical path to high-fidelity, stable avatar driving.

\end{abstract}

\section{Introduction}
\label{sec:intro}

Learning a high-fidelity, photorealistic virtual representation of a person has many applications for immersive experiences, such as virtual reality (VR), telepresence, video games and movies.
However, producing a person-specific avatar that captures fine details faithfully---\eg, cloth and loose-hair dynamics, subtle facial expressions, and characteristic motion patterns---typically requires extensive multi-view capture data.
Long-form captures often exhibit a pose--appearance ambiguity: the same skeletal pose can correspond to noticeably different appearances at different times.
This can happen, for example, when clothing is readjusted, hair settles differently, or wrinkles form in a new pattern.
A robust avatar model must simultaneously (i) represent high-frequency, view-dependent, and non-rigid effects, and (ii) remain temporally stable when the driving signal revisits similar poses.
Recent progress in 3D Gaussian Splatting (3DGS)~\cite{kerbl20233dgs} has made it an attractive representation for human avatars.
Its point-based structure provides the flexibility to model complex deformations, while enabling high-fidelity reconstructions that capture view-dependent effects and thin structures such as hair.
Yet, increasing model capacity introduces a second, closely related challenge: spurious correlations in the training data.
For instance, a particular hand configuration may be correlated with a specific wrinkle pattern on the trousers, even though it cannot cause that appearance change.
High-capacity models can exploit such correlations and memorize incidental details, causing erratic appearance changes and temporal flickering for novel pose sequences.

\begin{figure}[tb]
    \centering
    \includegraphics[trim=0cm 0cm 0cm 0.5cm,clip,width=1.0\linewidth]{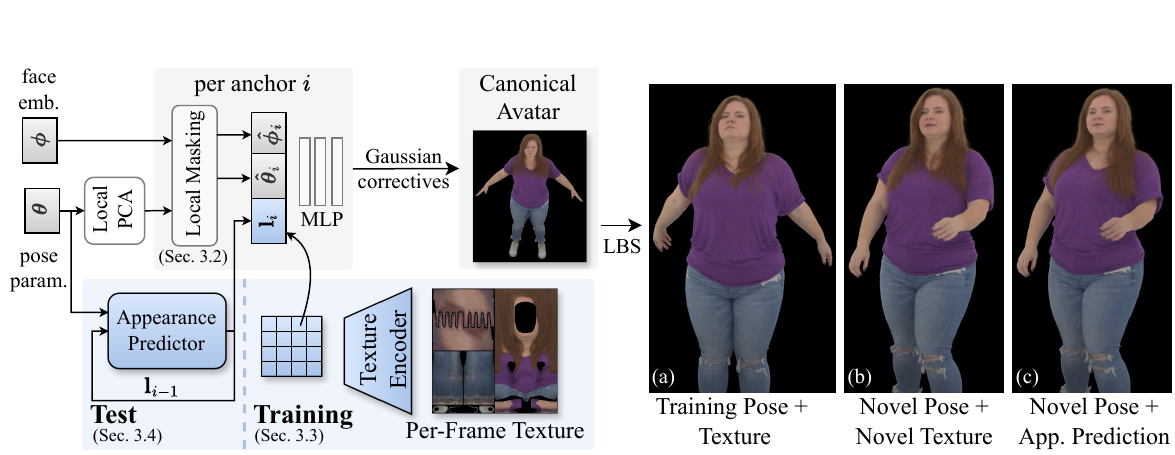}
    \caption{
    We represent the posed avatar with a hierarchical 3D Gaussian structure controlled by per-anchor spatial MLPs.
    Each anchor receives localized driving features via skinning-weight-based masking of pose parameters $\vect{\theta}$ (and face-region masking of $\vect{\phi}$), together with an appearance latent $\vect{l}_i$.
    During training, $\vect{l}_i$ is obtained by encoding a per-frame UV texture into a 2D feature map and sampling it at the anchor's UV coordinates; at test time, a transformer autoregressively predicts temporally smooth latents from a short pose history for stable driving.
    The encoder reconstructs training poses extremely well (a) and generalizes to novel poses and unseen textures (b), while our appearance predictor yields realistic appearances and smooth transitions on test sequences when textures are not available (c).
    }
    \label{fig:overview}
\end{figure}


We propose a 3DGS-based avatar model that addresses both \textbf{spurious pose correlations} and \textbf{pose\textendash\allowbreak appearance ambiguity} while maintaining high visual fidelity, dynamic effects, and smooth appearance transitions.
Our method is built on a hierarchical point representation~\cite{zhan2025spatialmlps}: a sparse set of anchors with spatially distributed MLPs, control points that guide positional displacement, and a dense set of 3D Gaussians that represent the remaining appearance attributes.
To reduce spurious long-range dependencies between unrelated pose parameters, we condition each spatial MLP only on \textbf{localized pose} information derived from skinning weights at its anchor location.
While this locality improves causal consistency, it also makes the pose--appearance ambiguity more pronounced: local pose parameters alone are often insufficient to determine the correct time-varying appearance.
To resolve this ambiguity, we introduce a per-frame appearance latent code learned during \textbf{training} via an \textbf{appearance encoder}.
The encoder takes as input a UV texture obtained by multi-view projection onto the template mesh, providing a compact, pose-aligned appearance observation without the need for computationally expensive per-frame mesh registration.
Because this UV texture is typically not available at \textbf{test time}, we train an additional transformer-based \textbf{appearance predictor} after avatar training to regress these encoder-produced latent codes.
Given a window of previous poses and the previous latent code, it predicts the next latent code, enabling temporally smooth and realistic appearance evolution during driving.

Combined, these components allow our model to fit challenging captures at exceptionally high fidelity while producing stable and plausible appearance transitions for novel pose sequences.
We validate our approach in quantitative and qualitative experiments on six extensive, high-quality captures, showing  significant improvements in temporal stability and adherence to the driving signal.

In summary, our contributions are:
\begin{itemize}
    \item A spatial-MLP-conditioned 3D Gaussian avatar model that combines localized pose conditioning with per-frame appearance latents to handle pose--appearance ambiguity and spurious pose correlations.
    \item A practical appearance encoding scheme that enables fast and stable learning of per-frame appearance latent codes during training.
    \item A transformer-based appearance predictor that generates temporally smooth appearance latents at test time from pose history.
\end{itemize}


\section{Preliminaries \& Related Work}
\label{sec:related_work}

As our work focuses on animatable, high-fidelity, and faithful reconstruction of person-specific avatars, we will not cover the large body of work on replay~\cite{weng2022humannerf,icsik2023humanrf,peng2023implicit,luiten2024dynamic3dgs} and monocular or single-image reconstruction~\cite{habermann2019livecap,jiang2022neuman,zhu2022registering,jiang2023instantavatar,hu2024gauhuman,hu2024gaussianavatar, deng2024ramavatar}.

\paragraph{Mesh-based:}

Mesh-based representations are widely used for modeling human avatars.
Early approaches either simulated clothing dynamics explicitly~\cite{guan2012drape} or synthesized novel views and motions via retrieval and interpolation from the captured dataset~\cite{xu2011videobased,casas20144dvideotextures}.
To reduce storage and improve fidelity for unseen poses, later works introduced neural networks to compute pose-dependent textures \cite{habermann2021real}.
Several methods further factorize the problem by modeling garments separately~\cite{xiang2021modelingclothing,xiang2022dressing}, through inverse physics~\cite{zheng2024physavatar}, and/or incorporate depth cues to improve drivability and robustness~\cite{xiang2023drivable}.
Mesh pipelines have also been combined with deferred shading and neural rendering modules~\cite{ma2021pixel}, or generative models~\cite{zhang2023getavatar}.
To better follow driving signals and reduce entanglement between motion and appearance, Bagautdinov et al.~\cite{bagautdinov2021drivingsignal} propose separating pose and appearance factors and enforcing disentanglement via mutual information minimization.

\paragraph{NeRF-based:}

Following NeRF~\cite{mildenhall2020nerf}, many avatar methods adopted volumetric rendering to better capture view-dependent effects and fine detail.
Most are template-based---either relying on a fitted template mesh~\cite{habermann2021real} or learning deformations of a canonical space~\cite{liu2021neuralactor,xu2021hnerf,peng2021animatablenerf,zheng2023avatarrex}---with template-free variants also explored~\cite{li2022tava}.
Although advancements in NeRFs improved its practicality via efficient encodings~\cite{muller2022instantngp} and factorized representations~\cite{chen2022tensorf}, real-time deployment---especially on low-power devices---remains challenging due to the multi-sample ray marching required for high quality, motivating more explicit alternatives.


\begin{figure}[!ht]
    \centering
    \includegraphics[width=0.33\linewidth]{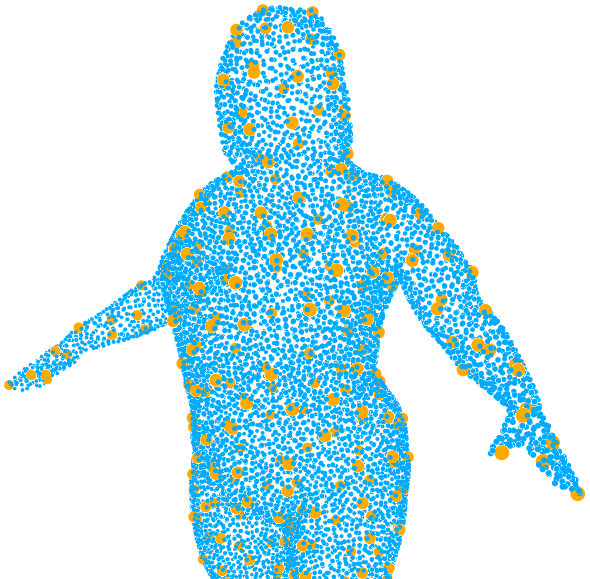}
    \includegraphics[width=0.38\linewidth]{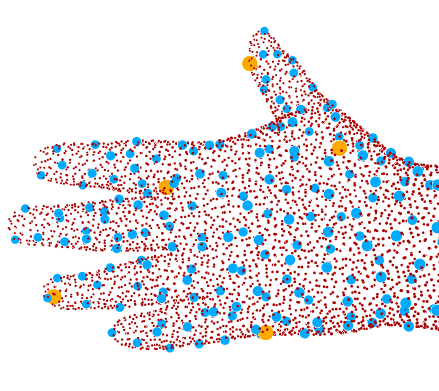}
    \caption{Following Zhan et al.~\cite{zhan2025spatialmlps}, we initialize a hierarchical point cloud on the template mesh, consisting of anchors/control points/Gaussians (300/10k/200k for our example, colored as orange/blue/red).
    Every anchor holds an MLP, whose outputs are interpolated by control points and Gaussians from the closest three anchors to calculate their positional displacement and Gaussian correctives.}
    \label{fig:hierarchie_viz}
\end{figure}

\paragraph{3DGS-based:}

Concurrently to NeRF-based methods, point-based avatar representations~\cite{ma2021pop,lin2022learningimplicittemplates} emerged, offering flexible modeling of complex deformations and convenient manipulation due to the independence of individual primitives.
3D Gaussian Splatting (3DGS)~\cite{kerbl20233dgs} largely superseded earlier point-based and many NeRF-based formulations by combining point-based flexibility with smooth spatial falloffs and favorable optimization behavior.
Building on 3DGS, early avatar methods focused primarily on geometric reconstruction and animation, often with limited pose-dependent appearance~\cite{lei2024gart,kocabas2024hugs}.
Subsequent work introduced stronger pose-conditioned deformation and shading models, e.g., via large MLPs~\cite{qian20243dgsavatar,moreau2024humangs} or CNN-based architectures~\cite{pang2024ash,jiang2025uvgaussians,chen2024meshavatar}.
Other directions target improved drivability and cloth behavior~\cite{zielonka2025drivable3dgs}, relightability~\cite{wang2025relightable}, or efficiency for mobile and VR rendering~\cite{iandola2025squeezeme,shao2024splattingavatar,chen2025taoavatar}.
Animatable Gaussians~\cite{li2024animatablegaussians}, predicts front--back UV maps with a single large CNN and achieves high visual quality, but falls short of real-time requirements.
To improve efficiency without sacrificing capacity, Zhan et al.~\cite{zhan2025spatialmlps} propose a hierarchical representation consisting of anchors, control points, and Gaussians, where each anchor holds a small spatially localized MLP (visualized in \cref{fig:hierarchie_viz}).
The MLP outputs are interpolated and combined with a per-Gaussian linear basis to produce Gaussian property correctives; positional updates are instead propagated from control points to enforce smooth deformations (see supplementary for details).
This design combines computationally cheap local MLPs with high representational power and well-behaved deformations, providing a compelling alternative to monolithic MLP- or CNN-based predictors.


\section{Method}
\label{sec:method}

This section details our avatar model and driving pipeline.
We begin with a high-level overview of the representation and conditioning strategy (\cref{subsec:method_overview}).
We then describe how we construct \emph{localized} pose features at each anchor to mitigate spurious pose--appearance correlations and reduce overfitting (\cref{subsec:method_localization}).
Next, we introduce our appearance modeling approach, including the per-frame latent learned during training and the encoder used to obtain it (\cref{subsec:method_app_modeling}).
Finally, we present the transformer-based appearance predictor used at test time to auto-regressively infer latents for driving.(\cref{subsec:method_app_prediction}).

\subsection{Overview}
\label{subsec:method_overview}

Given driving pose parameters $\vect{\theta}$~\cite{meta2025mhr} and face embeddings $\vect{\phi}$, our model outputs a posed avatar represented as a set of 3D Gaussians.
We follow the hierarchical representation of \cite{zhan2025spatialmlps}, using a sparse set of anchor points with spatially distributed MLPs that predict Gaussian parameters from per-anchor inputs (\cf \cref{fig:hierarchie_viz} for a visualization).
All points are initialized on the template mesh, receiving transferred skinning weights for linear blend skinning (LBS) and UV coordinates for texture-space lookups.
To reduce spurious long-range pose correlations, each anchor $i$ receives \textbf{localized} driving features: we apply skinning-weight-based \textbf{local masking} to obtain $\vect{\hat{\theta}}_i$, and mask $\vect{\phi}$ outside the face UV region to obtain $\vect{\hat{\phi}}_i$.
To further stabilize driving, we constrain poses using PCA computed from the training set \cite{li2024animatablegaussians}; importantly, we apply this transform during both training and test time.
Because a single global PCA entangles distant pose parameters, we perform \textbf{local PCA} separately over seven body regions.

Finally, we explicitly \textbf{model appearance} with per-frame local latent codes.
During training, we encode a per-frame UV texture---extracted via multi-view projection onto the template mesh---with a convolutional appearance encoder into a 2D feature map, and bilinearly sample it using each anchor's UV coordinates to obtain $\vect{l}_i$.
At test time, when UV textures are typically unavailable, we replace the encoder with a transformer \textbf{appearance predictor} that regresses the next latents from a short history of masked poses (and the previous latent), enabling smooth and stable appearance evolution during driving.
A full overview of our method can be seen in \cref{fig:overview}.

\paragraph{Losses.}

We train our model with a number of commonly used losses for human avatar reconstruction.
Specifically, we use a photometric $\ell_1$ loss $\lossL{1}$, a perceptual loss $\lossL{lpips}$~\cite{wang2004ssim}, an opacity regularizer $\lossL{opac}$ that encourages opaqueness in non-boundary regions, and a scale regularizer $\lossL{scale}$ to discourage overly large Gaussians.
Following \cite{zhan2025spatialmlps}, we enforce smooth deformations by penalizing differences between each control point's displacement and that of its five neighbors, yielding $\lossL{cpt}$.
Finally, we encourage a well-behaved and smooth appearance latent space via a Kullback--Leibler divergence penalty $\lossL{KL}$.
Additional training details are provided in the supplementary material.
The total loss is
\begin{align}
    \mathcal{L}= \lossL{1} + \lossterm{lpips} + \lossterm{opac} + \lossterm{scale} + \lossterm{cpt} + \lossterm{KL}.
\end{align}

\subsection{Localized Pose Parameters}
\label{subsec:method_localization}

\begin{figure}[tb]
  \centering
  \begin{subfigure}{0.55\linewidth}
  \centering
    \includegraphics[width=0.37\linewidth]{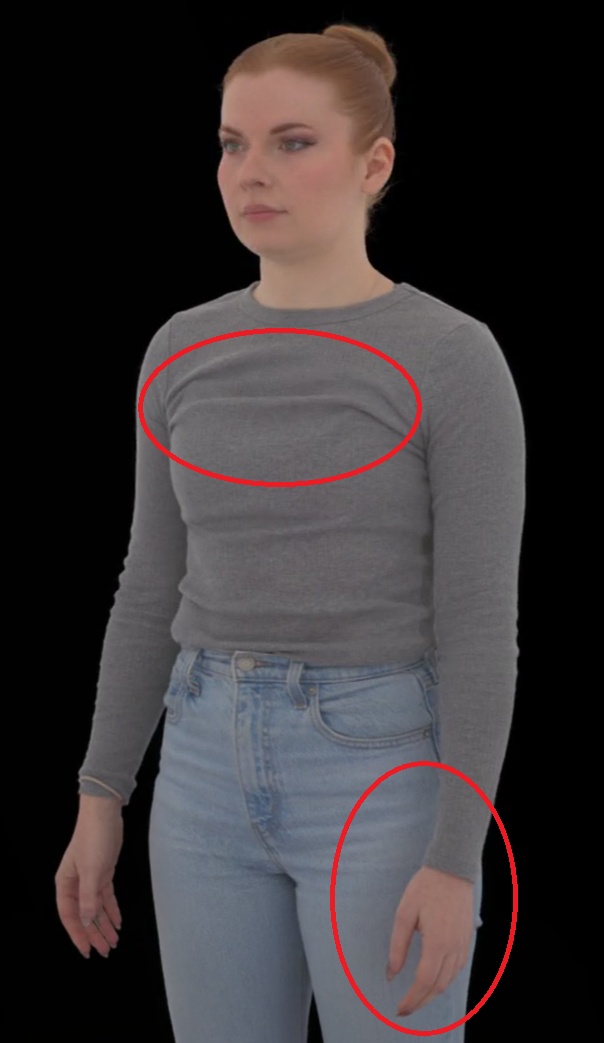}
    \includegraphics[width=0.3725\linewidth]{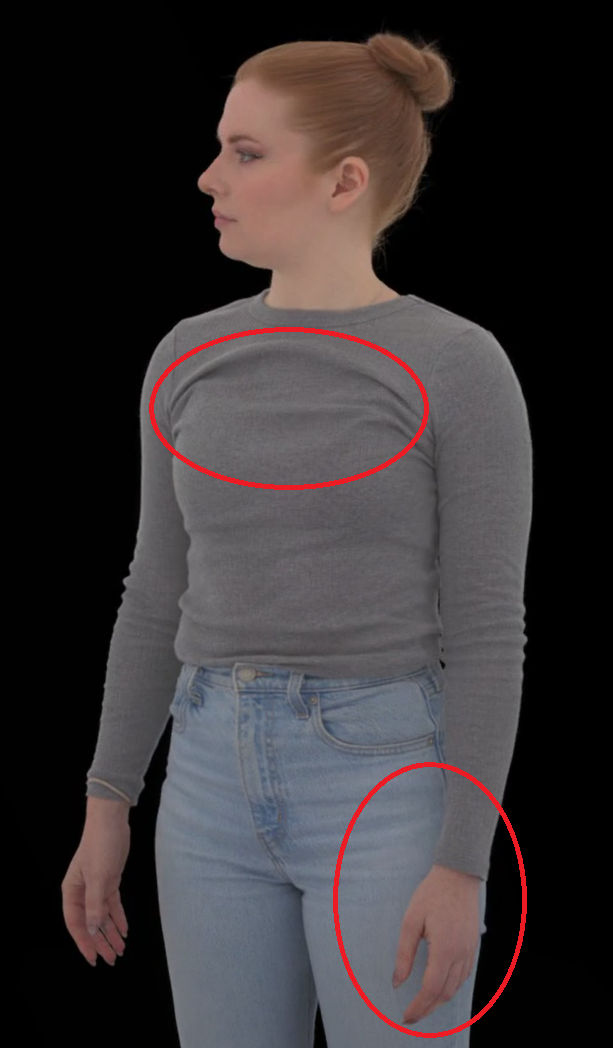}
    \caption{Spurious correlation of the neck parameter with unrelated regions (\eg, wrinkles and arm's shadow)}
    \label{fig:spurious_corr}
  \end{subfigure}
  \hfill
  \begin{subfigure}{0.41\linewidth}
  \centering
    \includegraphics[width=0.74\linewidth]{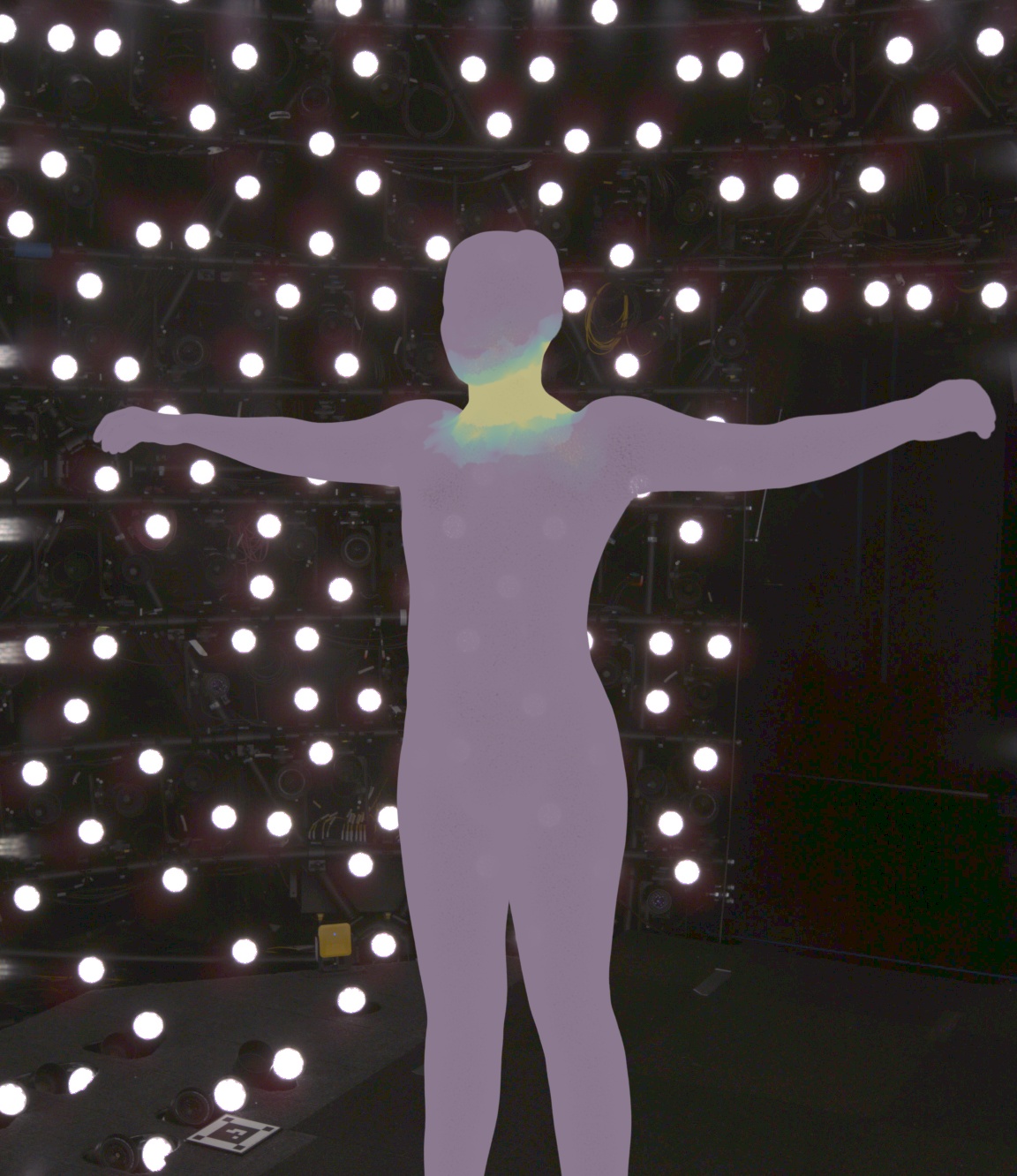}
    \caption{Region of influence for the neck twist pose parameter, visualized per Gaussian.}
    \label{fig:localization_mask_neck}
  \end{subfigure}
  \caption{(a) Providing pose parameter to every anchor leads to spurious correlations between unrelated regions. (b) We locally mask out pose parameters per anchor based on the skinning weights of the template mesh to restrict them to their local region.}
  \label{fig:localized_params}
\end{figure}

Conditioning every anchor on the full pose vector can induce spurious correlations between parameters at distant surface regions---\eg, a head rotation becoming correlated with a particular shirt wrinkle pattern (\cref{fig:spurious_corr}).
Such shortcuts are easy for high-capacity decoders to memorize and often manifest as abrupt appearance changes under novel driving motions.
Since non-rigid deformations and cloth dynamics are predominantly local, we introduce a pose-localization scheme that limits each anchor's receptive field in pose space.

\paragraph{Localized Masking.}

We restrict each pose parameter to the surface regions it can plausibly influence by masking the global pose vector $\vect{\theta}\in\mathbb{R}^P$ per anchor.
For an anchor $i$ on the template mesh, we trace the connection from its skinning weights to the pose parameters and construct a binary mask $\vect{m}^\theta_i{\in}\{0,1\}^P$, where $\vect{m}^\theta_i[j]{=}1$ iff any non-zero skinning weight affects $\vect{\theta}[j]$.
An example mask is shown in \cref{fig:localization_mask_neck}.
Similarly, we construct a mask $m^\phi_i{\in}\{0,1\}$ that is set if the anchor's texture coordinates lie inside the face region.
The anchor MLP then receives the localized pose and face embeddings
\begin{equation}
\vect{\hat{\theta}}_i = \vect{m}^\theta_i \odot \vect{\theta}, \quad \vect{\hat{\phi}}_i = m^\phi_i \vect{\phi},
\end{equation}
instead of the full $\vect{\theta}, \vect{\phi}$.
In practice, skinning-weight locality is well suited for tight clothing, but can be overly restrictive for looser garments whose motion may couple across larger regions.
To account for this, we dilate $\vect{m}^\theta_i$ once by propagating active entries to its neighboring anchors.
Finally, because finger pose parameters can have very small spatial support and may not cover any anchors, we reuse the wrist mask for the corresponding hand and finger parameters.

\paragraph{Localized PCA.}

Driving with poses far outside the training distribution may produces severe artifacts.
Following prior work~\cite{li2024animatablegaussians, zhan2025spatialmlps}, we therefore constrain poses using a PCA model fit to all training poses: we transform a pose into PCA space, clamp coefficients to $\pm2\sigma$, and apply the inverse transform.
Unlike previous approaches, we apply the same PCA transform and clamping during \emph{both} training and test time, avoiding a train--test distribution mismatch.

A drawback of global PCA is that it directly entangles distant pose parameters during training, which can reintroduce long-range correlations even in the presence of local masking.
To mitigate this, we perform \textbf{localized PCA}: we partition pose parameters into seven body-part groups and run PCA independently per group, using $5$ PCA components each.
The groups are torso+head (12 pose parameters), left/right leg (9 each), left/right arm (12 each), and left/right hand (27 each).
This preserves correlations within a body part while preventing global entanglement across unrelated regions.

\subsection{Appearance Modeling}
\label{subsec:method_app_modeling}

\begin{figure}[tb]
    \centering
    \begin{subfigure}{0.6\linewidth}
        \centering
        \begin{tabular}{ccc}
            \includegraphics[trim=0cm 0cm 0cm 3cm,clip,width=0.45\textwidth]{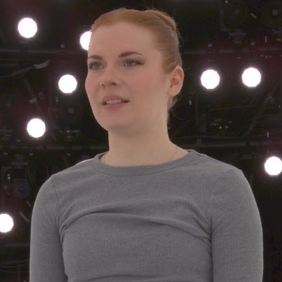} &
            \includegraphics[trim=0cm 0cm 0cm 3cm,clip,width=0.45\textwidth]{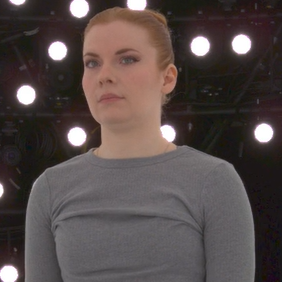} \\
            \includegraphics[trim=0cm 0cm 0cm 3cm,clip,width=0.45\textwidth]{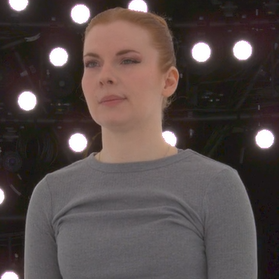} &
            \includegraphics[trim=0cm 0cm 0cm 3.1cm,clip,width=0.45\textwidth]{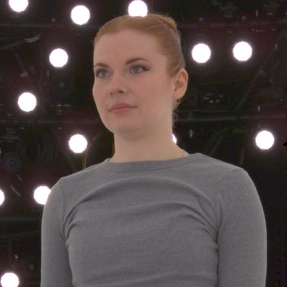}
        \end{tabular}
        \caption{Different appearance (folding) on very similar poses.}
        \label{fig:appearance_collage}
    \end{subfigure}
    \hfill
    \begin{subfigure}{0.38\linewidth}
        \centering
        \includegraphics[width=.9\linewidth]{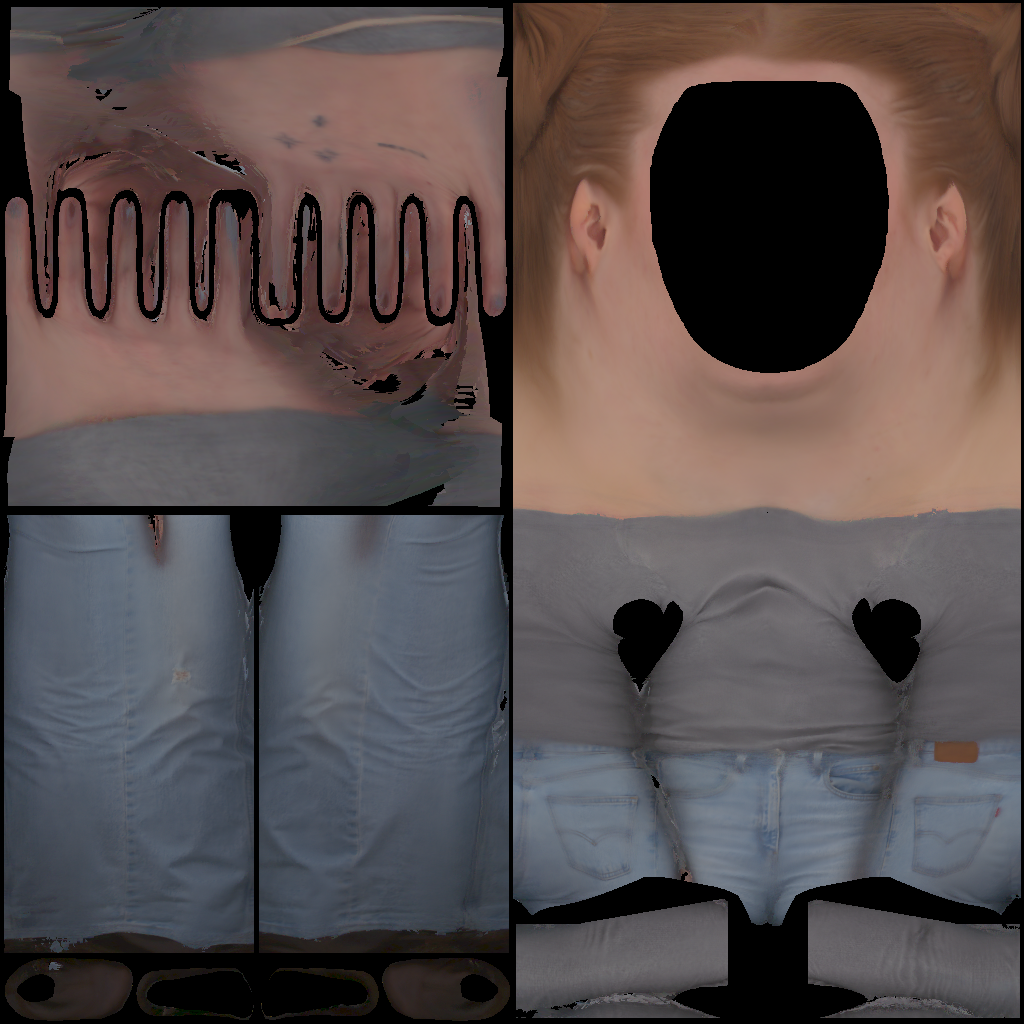}
        \caption{Input to the appearance encoder.}
        \label{fig:appearance_uv_texture}
    \end{subfigure}
    \caption{(a) Similar poses can exhibit vastly different appearances, leading to an ambiguous one-to-many mapping. Therefore, we model appearance through a texture encoder that receives (b) a multi-view projected UV texture as input (face region masked).}
    \label{fig:appearance}
\end{figure}

Long capture sequences---particularly with rigid garments or loose hair---often exhibit a pronounced pose--appearance ambiguity: nearly identical poses can correspond to noticeably different appearance (\eg, hair settling differently or new wrinkle configurations when returning to a neutral stance; see \cref{fig:appearance_collage}).
When an avatar is driven by pose alone, a high-capacity decoder can memorize these incidental correlations, leading to abrupt switching between appearance states.
To explicitly decouple pose from time-varying appearance, we introduce a learnable per-frame appearance latent.

A naïve per-frame codebook is data-inefficient, typically requires additional temporal regularization to yield a coherent latent space, and converges slowly.
Instead, we learn the latent through a convolutional texture encoder that takes as input a per-frame UV texture extracted by multi-view projection onto the template mesh (see \cref{fig:appearance_uv_texture} and the supplementary for details).
While this projection is imperfect due to small misalignments between the template mesh and the posed surface, it provides a sufficiently stable, pose-aligned reference that accelerates convergence and yields meaningful appearance codes without requiring computationally expensive mesh registration and texture fitting.
Since facial detail and expression should be explained by the face embedding $\vect{\phi}$, we mask the face region in the UV texture before encoding.


The encoder maps a $1024{\times}1024$ UV texture to a low-resolution feature map of size $32{\times}32{\times}N_l$.
For each anchor, we bilinearly sample this map at the anchor's UV coordinate to obtain a local latent code $\vect{l}_i$.
Compared to global latents, spatially varying codes better match the locality of non-rigid appearance changes, while bilinear sampling enforces smooth transitions across neighboring anchors.

\subsection{Appearance Prediction}
\label{subsec:method_app_prediction}

\begin{figure}[tb]
    \centering
    \includegraphics[trim=0.7cm 0.6cm 1.3cm 0cm,clip,width=0.75\linewidth]{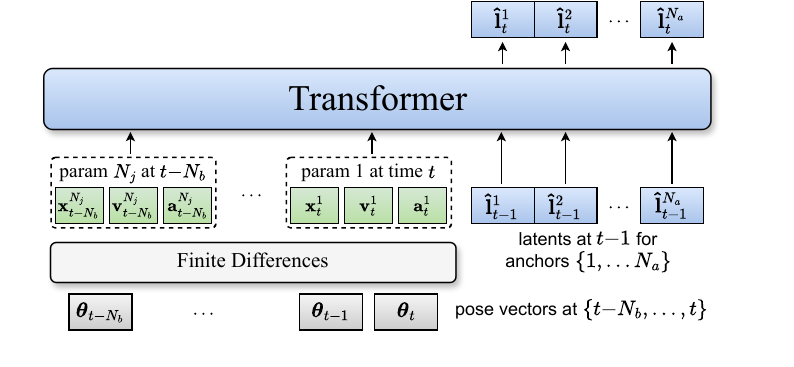}
    \caption{
    Architecture of our appearance predictor.
    The transformer predicts the next latents from the previous $N_b$ poses---value, plus their velocity and acceleration calculated via finite differences---and the previous latents.
    }
    \label{fig:motion_dynamics_model}
\end{figure}

To drive the avatar at test time without UV texture inputs, we learn an appearance predictor that infers the per-anchor appearance latents from motion history.
We model latent dynamics with a causal autoregressive transformer that takes the previous $N_b$ body poses and the previous-frame appearance codes as context, and predicts the appearance codes for the current frame (see \cref{fig:motion_dynamics_model}).
Unless stated otherwise, we use $N_b{=}10$, which provides sufficient temporal context while generalizing well to unseen sequences.


Naïvely concatenating all pose and latent dimensions into a single vector per frame led to overfitting in our experiments.
Instead, we represent the input as a set of tokens, treating each local pose parameter and each local appearance code as its own token.
For pose tokens, we use an extended representation consisting of \emph{value}, \emph{velocity}, and \emph{acceleration}, where velocity and acceleration are computed via finite differences.
We additionally normalize each training input sequence with respect to its last frame by removing that frame's global translation and rotation from all poses in the input sequence.


The model uses separate lightweight modality encoders to project pose and appearance latent codes into a common token space: a dedicated shallow MLP encoder for pose tokens and another for appearance tokens.
We add learnable positional embeddings before the modality encoders, and feed the resulting tokens through a stack of self-attention transformer blocks.
The output tokens represent the current frame's local appearance codes.


\paragraph{Training.}

At inference, the model must bootstrap without an existing historical context.
Initializing the context with zeros can create a train--test mismatch and destabilize autoregressive decoding.
To close this gap, we train with dual-context initialization: for each training sample we forward the model twice---(i) with ground-truth history appearance codes and (ii) with zeroed codes---and sum the losses from both predictions.
This enables recovery from zero initialization while maintaining accuracy when a valid history exists.
Autoregressive decoding can also suffer from error accumulation over long sequences.
To improve robustness, we explicitly unroll the autoregressive predictions over the input window during training---\ie, feeding back the model's prediction at time $t{-}1$ as input for time $t$---and we accumulate losses across the unrolled window.
Finally, we apply PCA with clamping (as in \cref{subsec:method_localization}) to the predicted appearance codes during both training and testing, ensuring outputs remain within the training latent distribution and reducing collapse and visual artifacts.

\paragraph{Losses.}

We supervise appearance prediction with an $\ell_1$ loss on the latent codes for both the ground-truth- and zero-initialized prediction.
To encourage temporal stability, we additionally apply $\ell_1$ penalties to finite-difference derivatives of the velocity, acceleration, and jerk of the latent code output.
All loss terms are weighted equally (weight $1$).

\section{Evaluation}

We evaluate our method on six extensive long-form dome captures spanning a range of difficulty and dynamic appearance effects.
We compare against a re-implementation of Zhan et al.~\cite{zhan2025spatialmlps}, augmented with our localized face embedding and localized hand/finger pose conditioning to allow for a fair comparison; we denote this baseline as \emph{MMLPs$^\dagger$}.
Additionally, we compare against a convolutional decoder baseline similar to a non-relightable version of the Relightable Full-Body Gaussian Codec Avatar (RFGCA) model from Want et al.~\cite{wang2025relightable}---denoted as \emph{nRFGCA\textsuperscript{$\dagger$}}.
All models are implemented in PyTorch and use \emph{gsplat}~\cite{ye2025gsplat} for differentiable 3D Gaussian rendering.

\paragraph{Dataset.}

Our dataset comprises six dome captures, each with ${\sim}$30k training frames at ${\sim}$30\,fps and ${\sim}$250 synchronized training cameras, split evenly between full-body, upper-body, and lower-body views.
Each capture contains both instructed motion sequences and less scripted segments.
All images (with corresponding foreground segmentation masks) for training and evaluation are rendered at a resolution of $1152\times1332$ (4${\times}$ downscaled from the captured images).
For evaluation, we hold out ${\sim}$3k frames and 6 cameras during training (four full-body cameras, plus one frontal upper-body and one frontal lower-body camera).


\paragraph{Training Details.}

We train and evaluate all methods on 8 NVIDIA H200 GPUs with a total batch size of 64 for 100k iterations, resulting in 8--10 hours of training time.
Following \cite{zhan2025spatialmlps}, \emph{Ours} and \emph{MMLPs$^\dagger$} use 300/10k/200k points for anchors/control points/Gaussians.
Our appearance model uses $N_l{=}16$ latent channels.
We optimize with AdamW using $\epsilon{=}10^{-15}$, $\beta_1{=}0.9$, and $\beta_2{=}0.999$, and apply weight decay only to the spatial MLPs and the appearance encoder, with $\lambda{=}0.001$.
The loss weights are set to $\losslam{lpips}{=}0.1,\losslam{lpips}{=}0.1,\losslam{opac}{=}0.5,\losslam{scale}{=}1.0$,
$\losslam{cpt}{=}0.5,\losslam{KL}{=}10^{-6}$.


\subsection{Results}

We present quantitative and qualitative results on all six captures.
Our evaluation focuses on (i) the overfitting capabilities of each method on the training data, (ii) the ability of our encoder to generalize to unseen textures, (iii) temporal stability and visual plausibility of our appearance predictor compared to the baselines \wrt spurious correlations and pose--appearance ambiguity.
We strongly encourage the reader to watch the supplementary videos, as the improvements in temporal stability and motion dynamics plausibility are more easily observable in videos than in still images or aggregate metrics.


\subsubsection{Quantitative.}

\begin{table}[tb]
    \centering
    \caption{
    Image metrics on all compared methods.
    As appearance is inherently ambiguous and drifts over time, we initialize the appearance predictor for Ours with the encoder output.
    Additionally, we show results for re-initializing every 30 frames, and for using the encoder during driving.
    }
    \setlength{\tabcolsep}{3pt}
\begin{tabular}{lccccccr}
 & \multicolumn{3}{c}{Test} & \multicolumn{3}{c}{Train}\\
 \cmidrule(lr){2-4} \cmidrule(lr){5-7}
 & PSNR\textsuperscript{$\uparrow$} & SSIM\textsuperscript{$\uparrow$} & LPIPS\textsuperscript{$\downarrow$} & PSNR\textsuperscript{$\uparrow$} & SSIM\textsuperscript{$\uparrow$} & LPIPS\textsuperscript{$\downarrow$} & \#Params \\ \hline
Ours & \cellcolor{best!40}32.99 & \cellcolor{best!40}0.939 & \cellcolor{best!40}0.063 & \cellcolor{best!40}36.21 & \cellcolor{best!40}0.950 & \cellcolor{best!40}0.053 & 192.1M\\
MMLPs\textsuperscript{$\dagger$} & \cellcolor{best!00}32.34 & \cellcolor{best!00}0.937 & \cellcolor{best!00}0.066 & \cellcolor{best!15}36.13 & \cellcolor{best!40}0.950 & \cellcolor{best!15}0.054 & 187.2M\\
nRFGCA\textsuperscript{$\dagger$} & \cellcolor{best!15}32.74 & \cellcolor{best!40}0.939 & \cellcolor{best!15}0.065 & \cellcolor{best!00}35.94 & \cellcolor{best!40}0.950 & \cellcolor{best!00}0.055 & 155.9M\\
\hline
Ours (re-init. 30) & \cellcolor{best!00}33.85 & \cellcolor{best!00}0.942 & \cellcolor{best!00}0.060 & -- & -- & -- & 192.1M \\
Ours (encoder) & \cellcolor{best!00}34.78 & \cellcolor{best!00}0.946 & \cellcolor{best!00}0.056 & -- & -- & -- & 189.8M
\end{tabular}
    \label{tab:results}
\end{table}

We report standard reconstruction metrics, including PSNR, SSIM~\cite{wang2004ssim}, and the perceptual similarity metric LPIPS~\cite{zhang2018perceptual}.
\cref{tab:results} summarizes training- and test-set performance for all compared methods.
On the training set, our method achieves the best scores, indicating that the added structure (localized conditioning and appearance latents) enables high-quality fitting of the training data.
On the test set, our full pipeline with the appearance predictor performs best overall.
Because appearance is inherently ambiguous and drifts over time, we expect the appearance immediately after initialization to match the ground truth closest.
Accordingly, we report results with the predictor re-initialized every 30 frames, which shows larger improvements over this interval.
Additionally, we report a test-time upper bound using the appearance encoder, which isolates the encoder and highlights its ability to generalize to novel poses and previously unseen textures.
We also find that temporal stability under driving improves substantially in practice, as demonstrated in the qualitative results and supplementary videos.
Finally, we further ablate our design choices in \cref{tab:results_ablation}.
Applying localized pose masking and localized PCA slightly reduces training-set fit, but improves generalization when driving with novel poses.


\begin{table}[tb]
    \centering
    \small
    \caption{Ablation of our components.}
    \setlength{\tabcolsep}{3pt}
\begin{tabular}{lcccccc}
 & \multicolumn{3}{c}{Test} & \multicolumn{3}{c}{Train} \\
 \cmidrule(lr){2-4} \cmidrule(lr){5-7}
 & PSNR\textsuperscript{$\uparrow$} & SSIM\textsuperscript{$\uparrow$} & LPIPS\textsuperscript{$\downarrow$} & PSNR\textsuperscript{$\uparrow$} & SSIM\textsuperscript{$\uparrow$} & LPIPS\textsuperscript{$\downarrow$} \\ \hline
Ours & \cellcolor{best!40}32.99 & \cellcolor{best!40}0.939 & \cellcolor{best!40}0.063 & \cellcolor{best!00}36.21 & \cellcolor{best!00}0.950 & \cellcolor{best!00}0.053 \\
Ours w/o local pca & \cellcolor{best!00}32.86 & \cellcolor{best!00}0.938 & \cellcolor{best!00}0.064 & \cellcolor{best!15}36.23 & \cellcolor{best!15}0.950 & \cellcolor{best!15}0.053 \\
Ours w/o local pose\textsubscript{\phantom{$\downarrow$}} & \cellcolor{best!15}32.87 & \cellcolor{best!15}0.938 & \cellcolor{best!15}0.064 & \cellcolor{best!40}36.29 & \cellcolor{best!40}0.951 & \cellcolor{best!40}0.052 \\
\end{tabular}
    \label{tab:results_ablation}
\end{table}


\subsubsection{Qualitative.}

\begin{figure}[!ht]
    \centering
    \begin{subfigure}{0.515\linewidth}
        \centering
        \setlength{\tabcolsep}{1pt}
        \begin{tabular}{ccccccc}
            & Frame $t$ & Frame $t{+}1$ & $\ell_1$ error \\
            \rotatebox[origin=l]{90}{\ \ \ \ Neck Twist} &
            \includegraphics[width=0.3\textwidth]{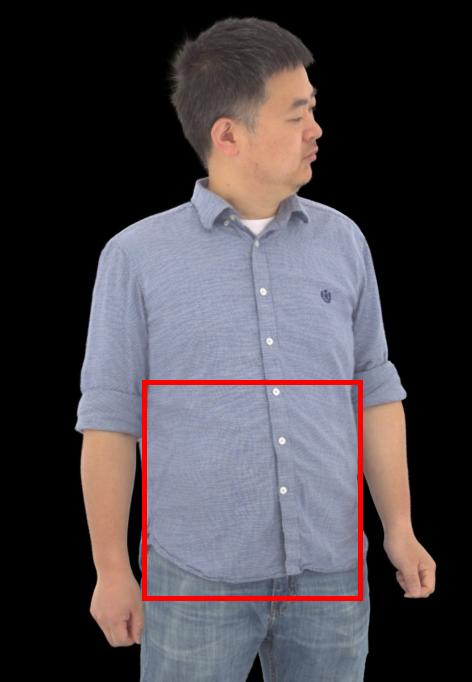} &
            \includegraphics[width=0.3\textwidth]{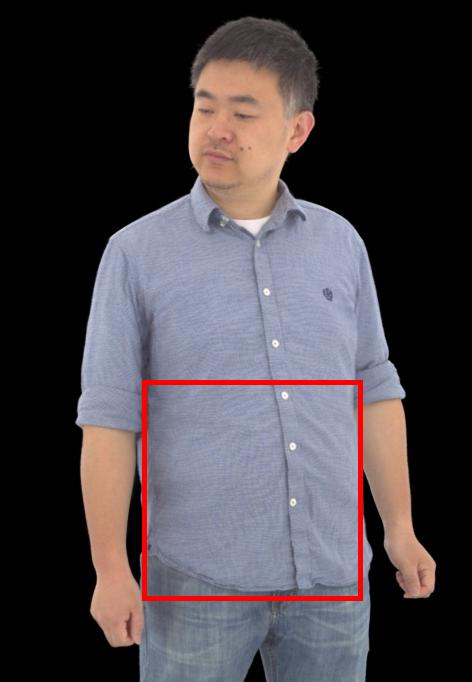} &
            \includegraphics[width=0.3\textwidth]{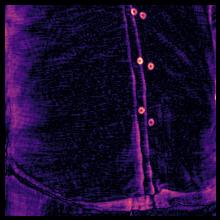} \\
            \rotatebox[origin=l]{90}{\ \ Wrist Rotate} &
            \includegraphics[width=0.3\textwidth]{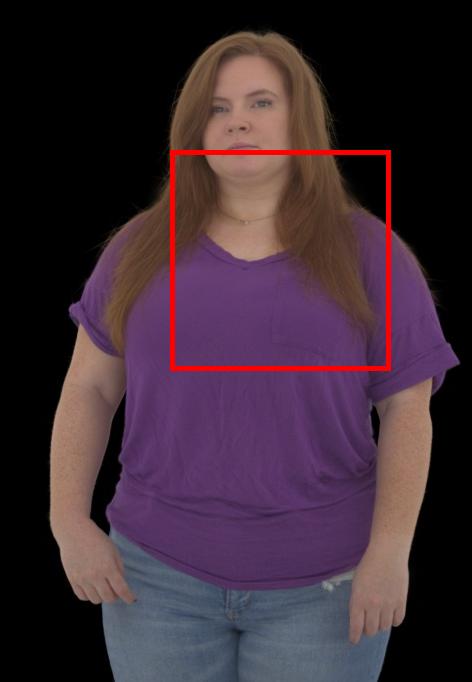} &
            \includegraphics[width=0.3\textwidth]{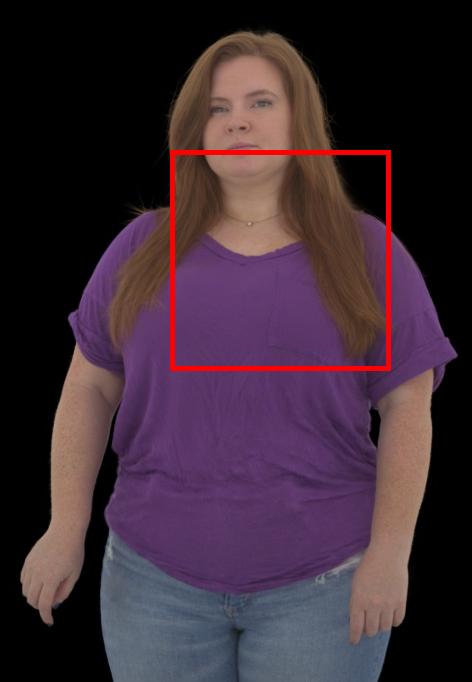} &
            \includegraphics[width=0.3\textwidth]{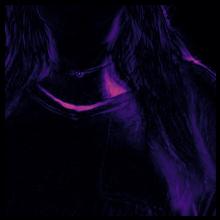}
        \end{tabular}
        \caption{MMLPs\textsuperscript{$\dagger$}}
    \end{subfigure}
    \hfill
    \begin{subfigure}{0.47\linewidth}
        \centering
        \setlength{\tabcolsep}{1pt}
        \begin{tabular}{cccccc}
            Frame $t$ & Frame $t{+}1$ & $\ell_1$ error \\
            \includegraphics[width=0.33\textwidth]{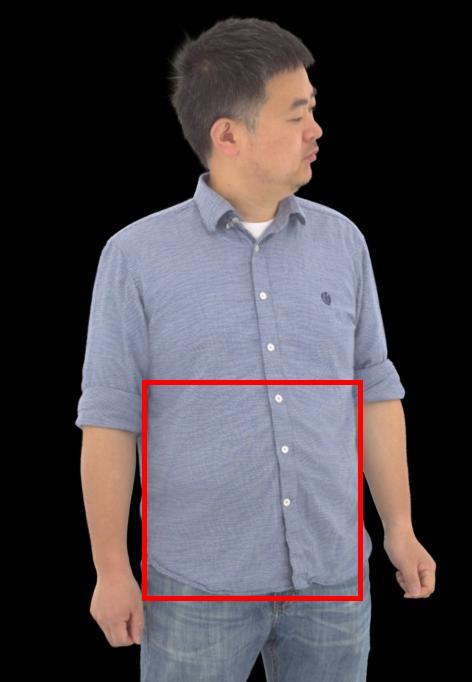} &
            \includegraphics[width=0.33\textwidth]{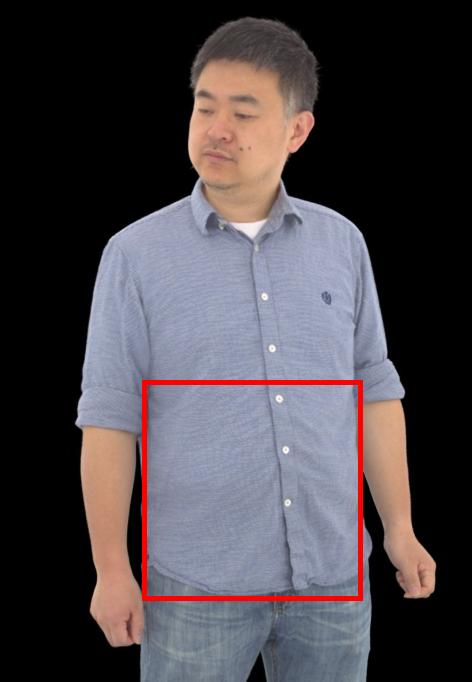} &
            \includegraphics[width=0.33\textwidth]{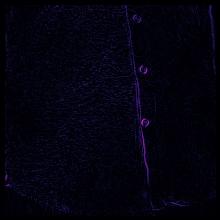} \\            \includegraphics[width=0.33\textwidth]{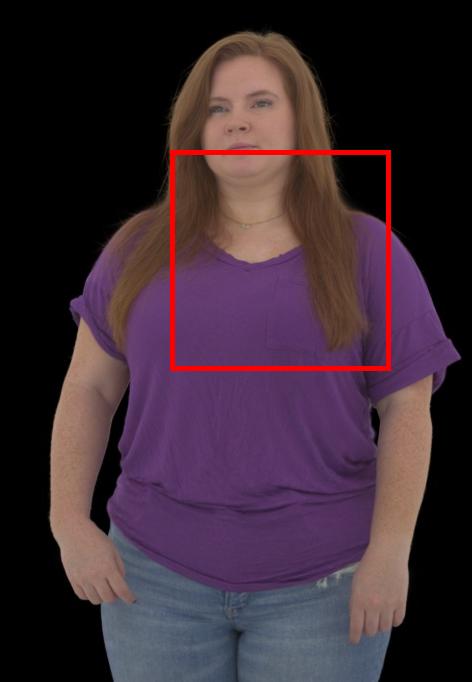} &
            \includegraphics[width=0.33\textwidth]{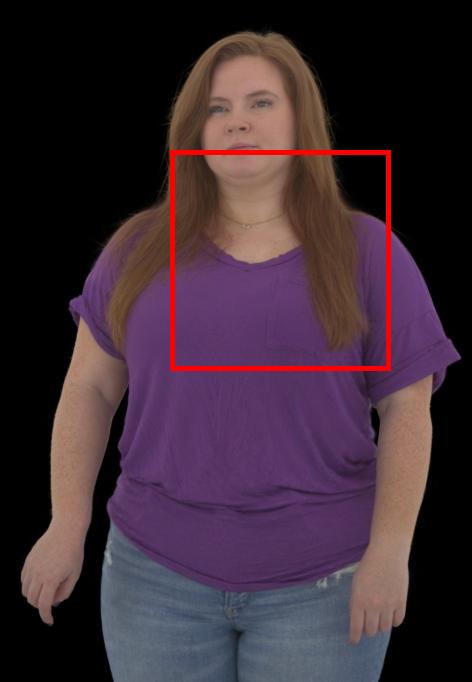} &
            \includegraphics[width=0.33\textwidth]{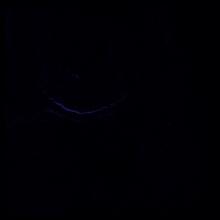}
        \end{tabular}
        \caption{Ours}
    \end{subfigure}
    \caption{Qualitative comparison of \textbf{global} (MMLPs\textsuperscript{$\dagger$}) versus \textbf{localized} (Ours) pose conditioning.
    We manually perturb a single pose parameter (top: neck twist; bottom: wrist rotate) between two frames: global conditioning induces large, non-local deformations and spurious appearance changes (visualized as $\ell_1$ error in the cutouts), whereas localized conditioning confines the effect to local, physically plausible motion.}
    \label{fig:localpose_collage_new}
\end{figure}

\cref{fig:localpose_collage_new} illustrates how our localized pose conditioning reduces spurious long-range correlations.
\emph{MMLPs$^\dagger$} conditions every anchor on the full pose vector, allowing the decoder to exploit incidental correlations in the training data; as a result, distant pose changes can spuriously modulate unrelated appearance, such as arm shadows, shirt wrinkles, or hair configuration.
In contrast, our approach supplies pose parameters only locally and applies localized PCA consistently at both training and test time, which suppresses unrealistic long-range effects.
Across captures, we observe that \emph{MMLPs$^\dagger$} is particularly prone to unstable shading and deformation due to the test-time-only global PCA, while access to global pose parameters further encourages spurious correlations.


\begin{figure}[tb]
    \centering
    \setlength{\tabcolsep}{1pt}
    \begin{tabular}{cccccccccc}
        \rotatebox[origin=l]{90}{\ \ \ \ \ \ \ \ \ \ Frame $t$} &
        \includegraphics[height=0.3\linewidth]{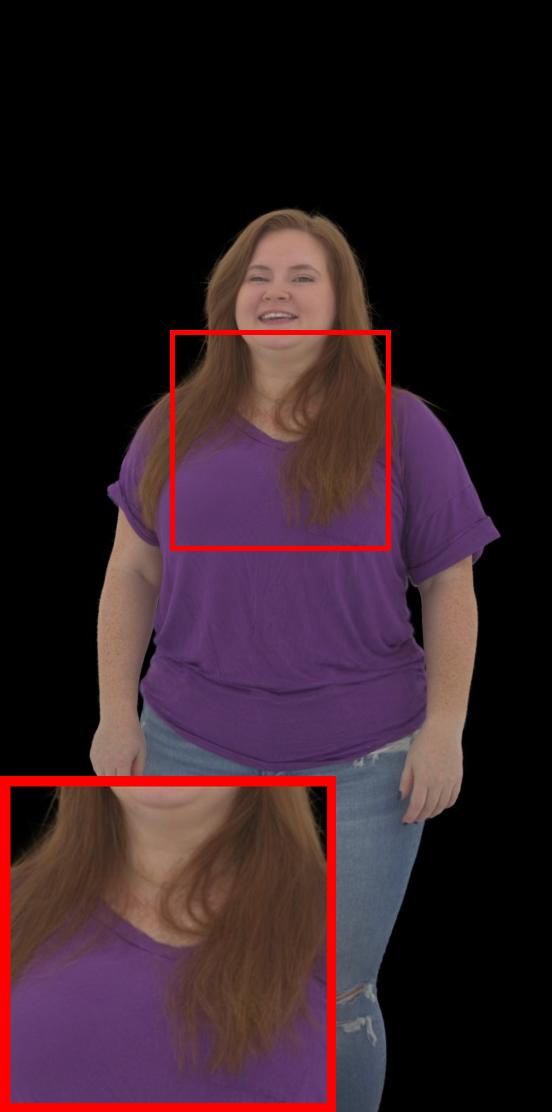} &
        \includegraphics[height=0.3\linewidth]{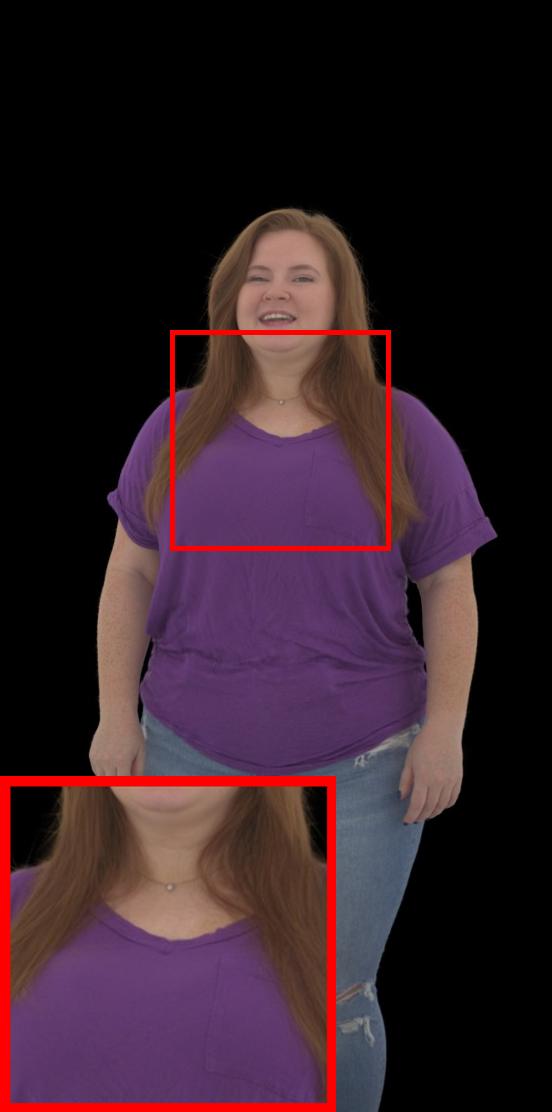} &
        \includegraphics[height=0.3\linewidth]{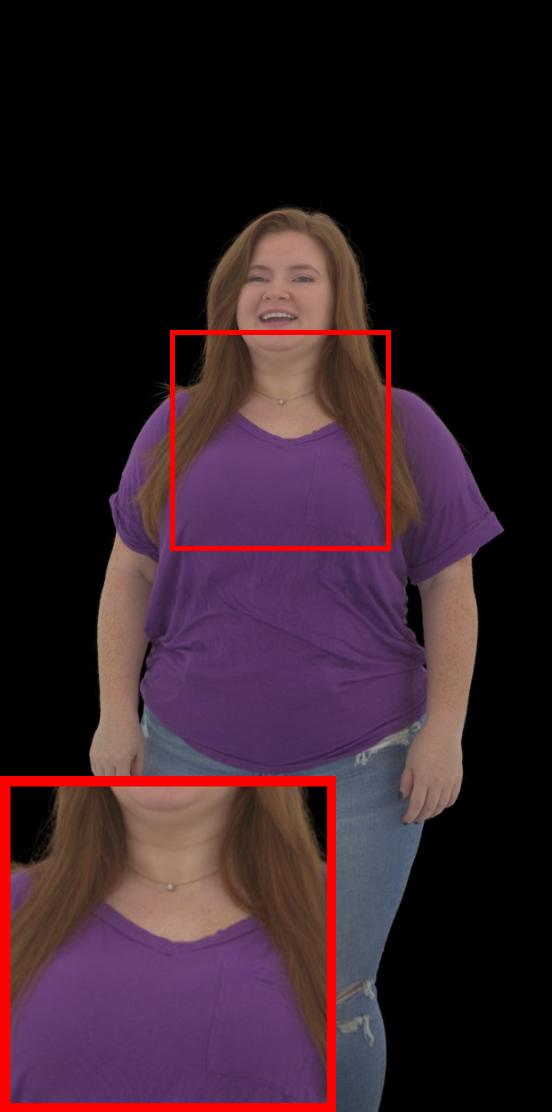} &&&&
        \includegraphics[height=0.3\linewidth]{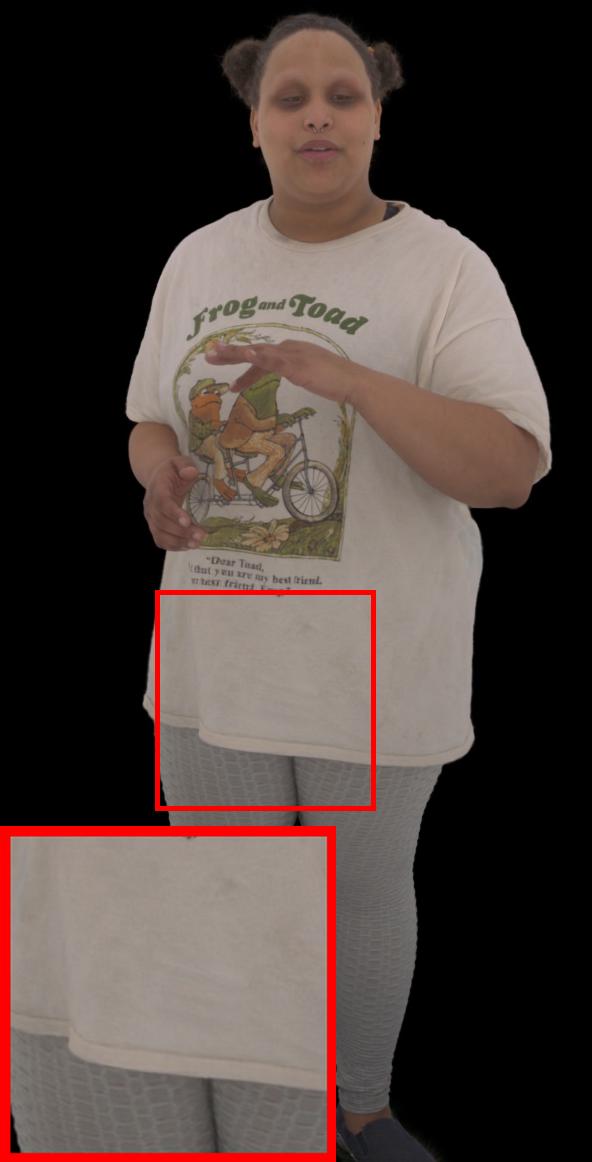} &
        \includegraphics[height=0.3\linewidth]{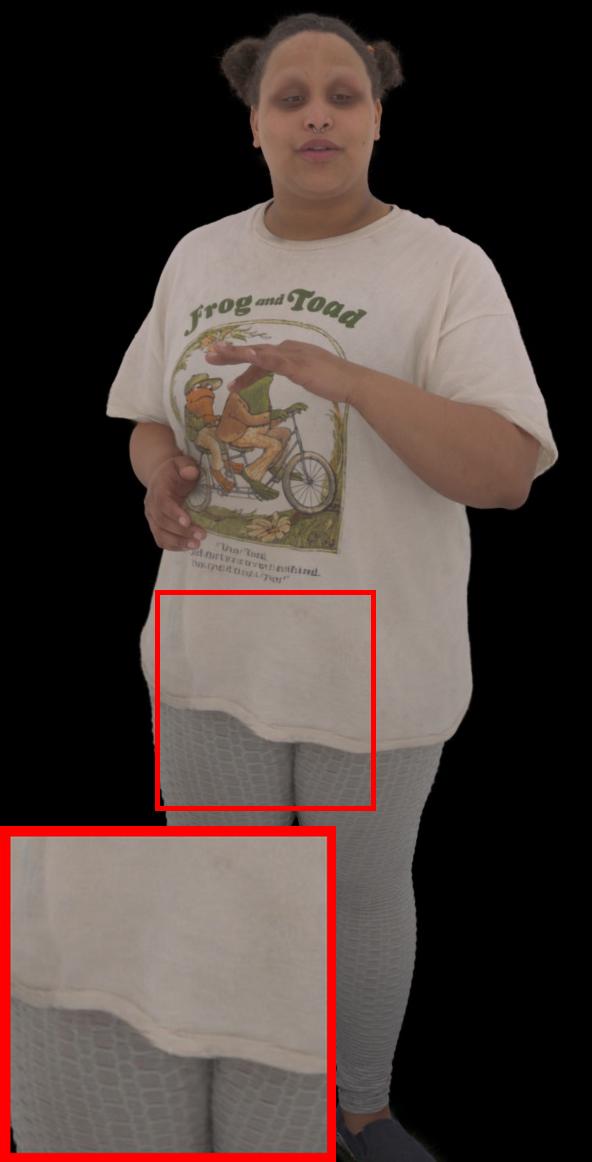} &
        \includegraphics[height=0.3\linewidth]{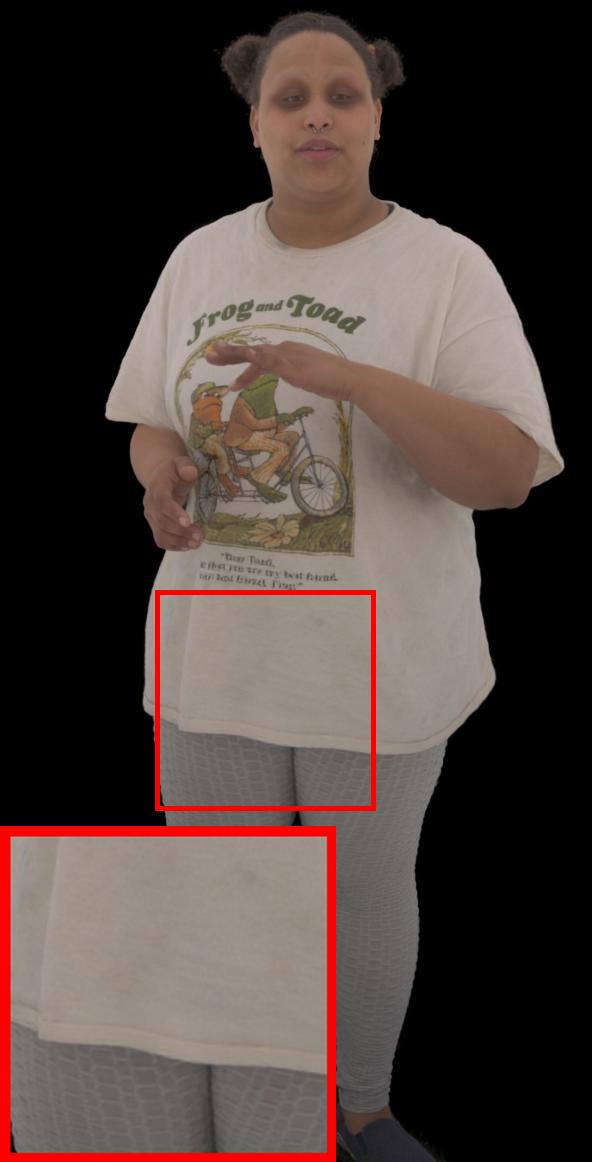} \\
        \rotatebox[origin=l]{90}{\ Frame $t{+}\delta$ ($\delta < 200$ms)} &
        \includegraphics[height=0.3\linewidth]{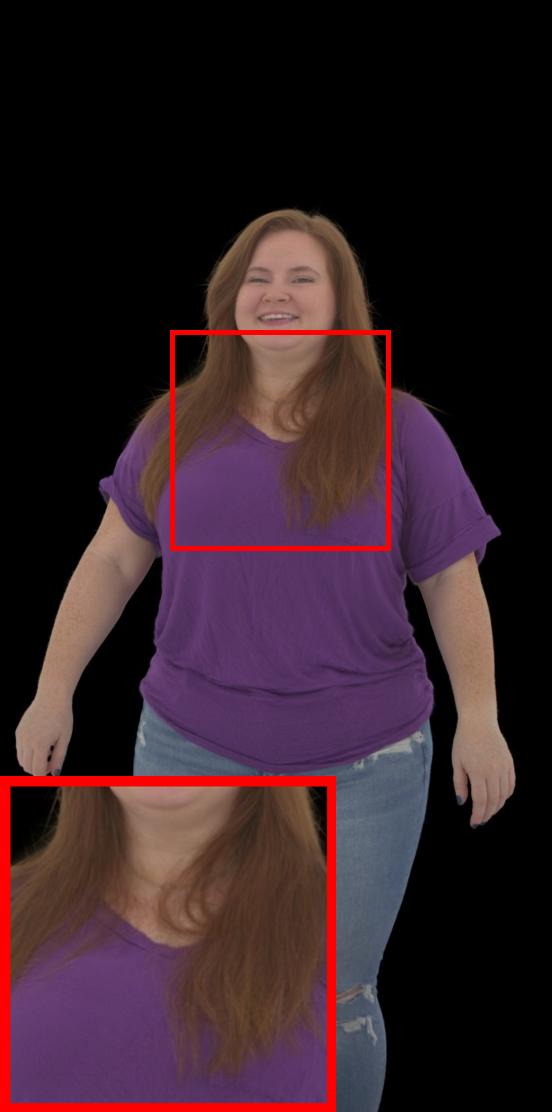} &
        \includegraphics[height=0.3\linewidth]{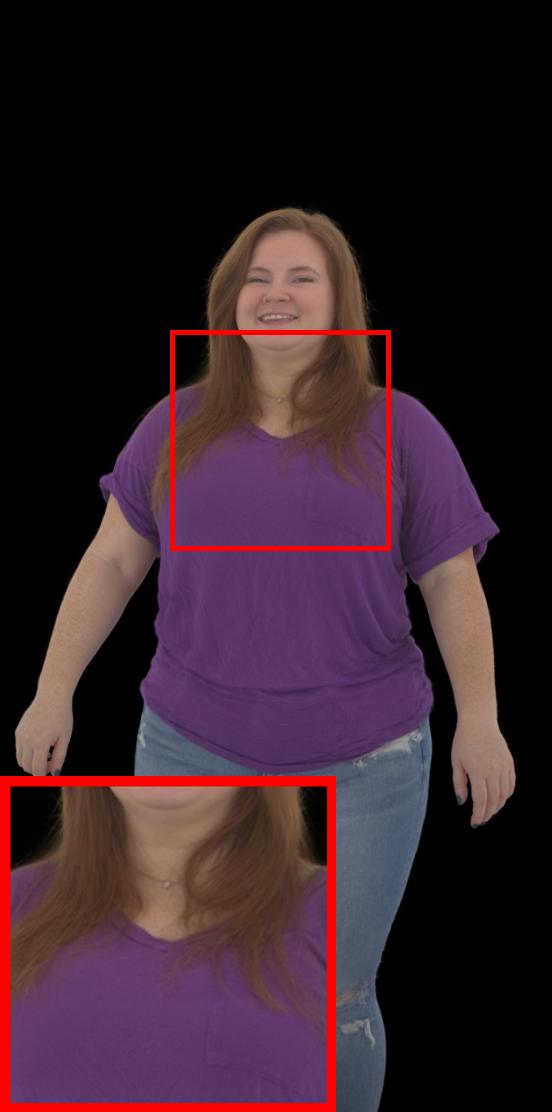} &
        \includegraphics[height=0.3\linewidth]{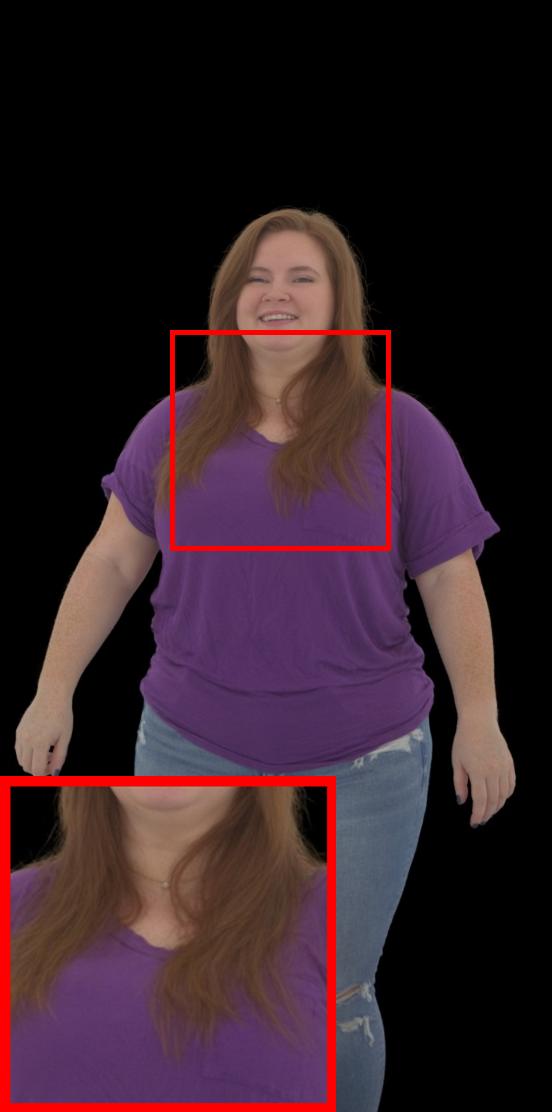} &&&&
        \includegraphics[height=0.3\linewidth]{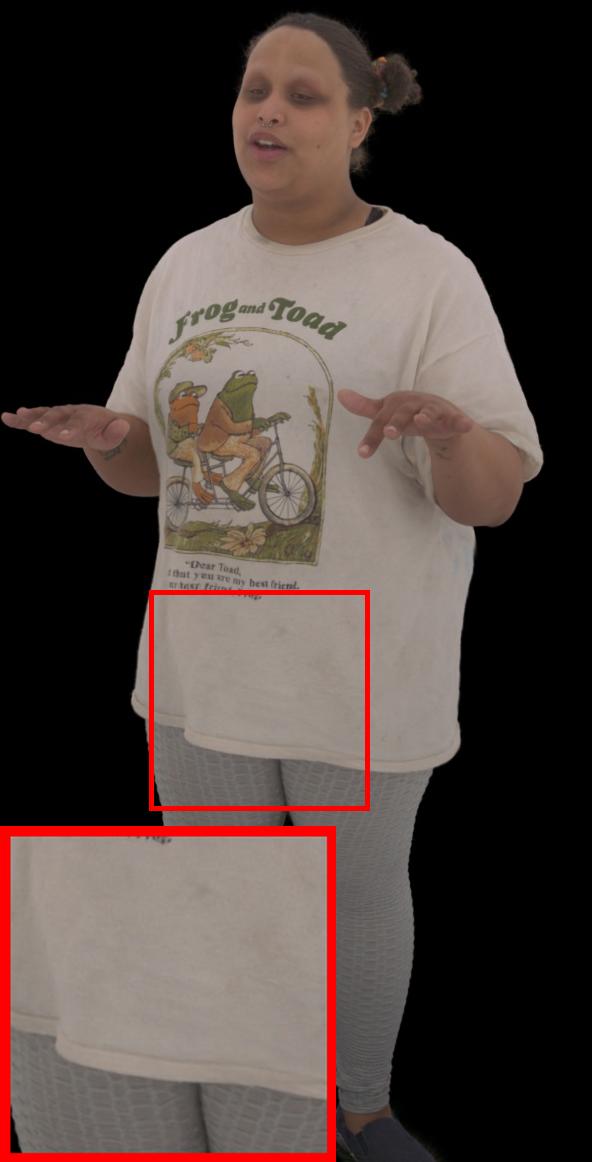} &
        \includegraphics[height=0.3\linewidth]{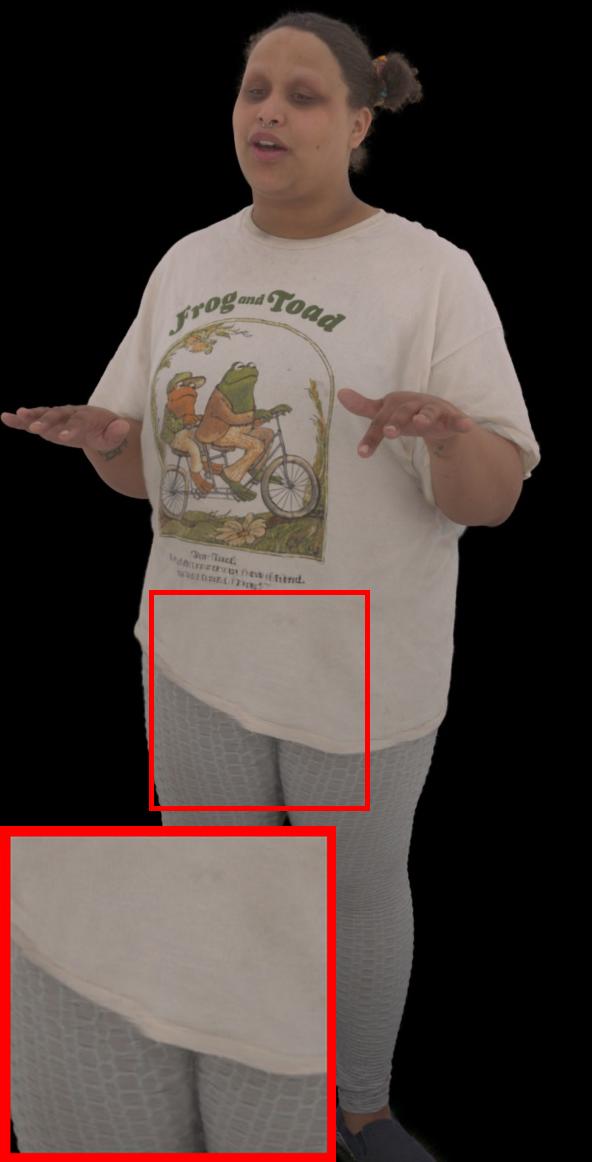} &
        \includegraphics[height=0.3\linewidth]{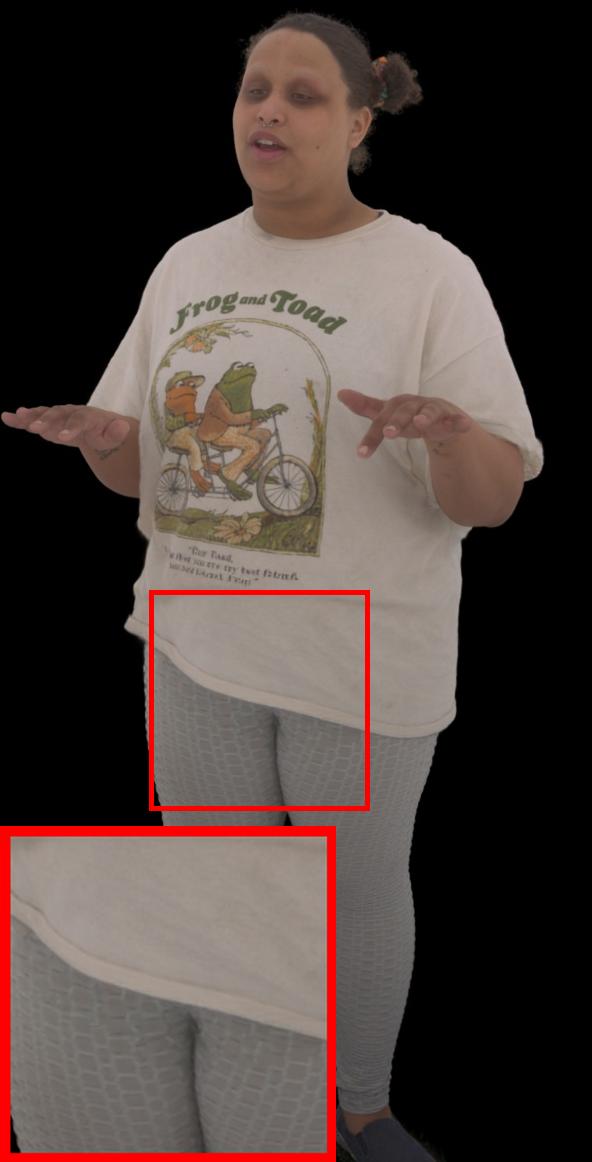} \\
        \rotatebox[origin=l]{90}{\ \ \ $\ell_1$ error} &
        \includegraphics[height=0.15\linewidth]{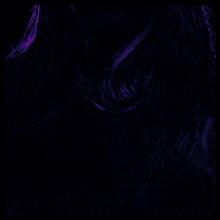} &
        \includegraphics[height=0.15\linewidth]{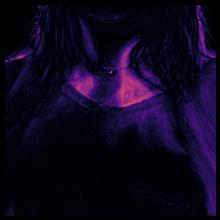} &
        \includegraphics[height=0.15\linewidth]{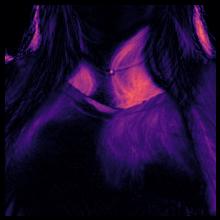} &&&&
        \includegraphics[height=0.15\linewidth]{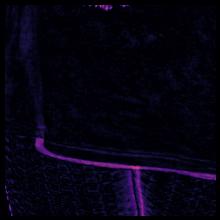} &
        \includegraphics[height=0.15\linewidth]{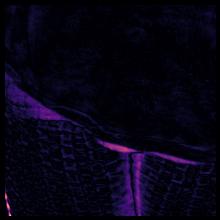} &
        \includegraphics[height=0.15\linewidth]{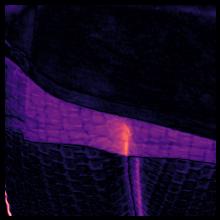} \\
        & Ours & MMLPs\textsuperscript{$\dagger$} & nRFGCA\textsuperscript{$\dagger$} &&&& Ours & MMLPs\textsuperscript{$\dagger$} & nRFGCA\textsuperscript{$\dagger$}
    \end{tabular}
    \caption{
    Two temporally adjacent frames from a test sequence and the corresponding cutout $\ell_1$ error against the previous frame.
    Our method remains temporally stable due to explicit pose--appearance disentanglement and the learned appearance predictor, while nRFGCA$^\dagger$ and MMLPs$^\dagger$ exhibit abrupt, implausible appearance changes.}
    \label{fig:temporal_stability_collage}
\end{figure}

To assess temporal stability, \cref{fig:temporal_stability_collage} shows two temporally adjacent frames from a test sequence.
Both \emph{nRFGCA$^\dagger$} and \emph{MMLPs$^\dagger$} can exhibit noticeable flicker on challenging captures, with abrupt and implausible appearance changes over very short time spans.
Our method, driven by the appearance predictor, produces substantially more stable results while still allowing realistic, pose-dependent local appearance evolution from a short motion history.
This effect is most clearly visible in the supplementary videos.


\begin{figure}[!ht]
    \centering
    \begin{tabular}{cccc}
        \includegraphics[width=0.24\textwidth]{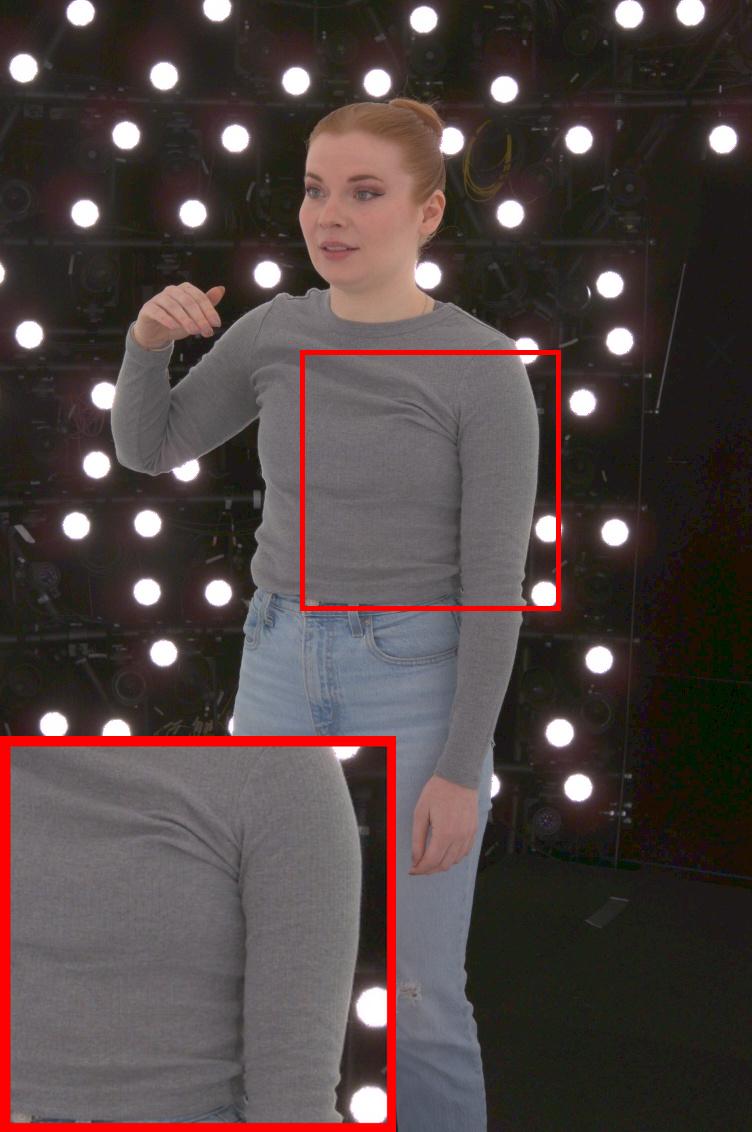} &
        \includegraphics[width=0.24\textwidth]{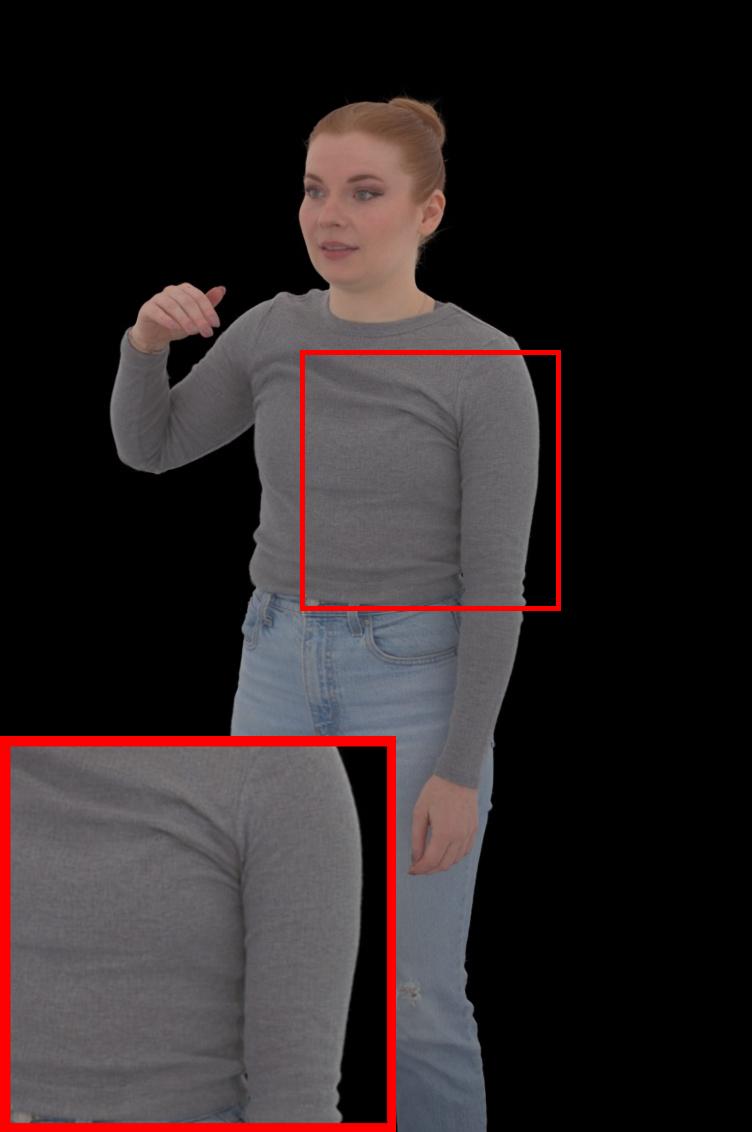} &
        \includegraphics[width=0.24\textwidth]{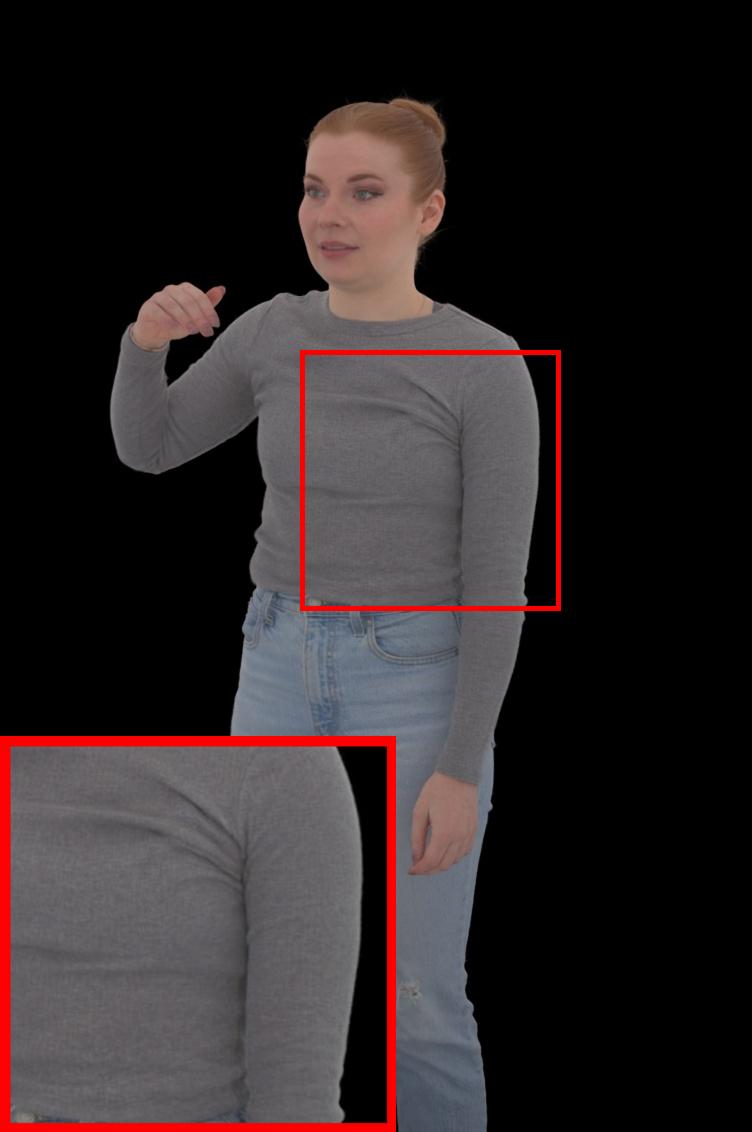} &
        \includegraphics[width=0.24\textwidth]{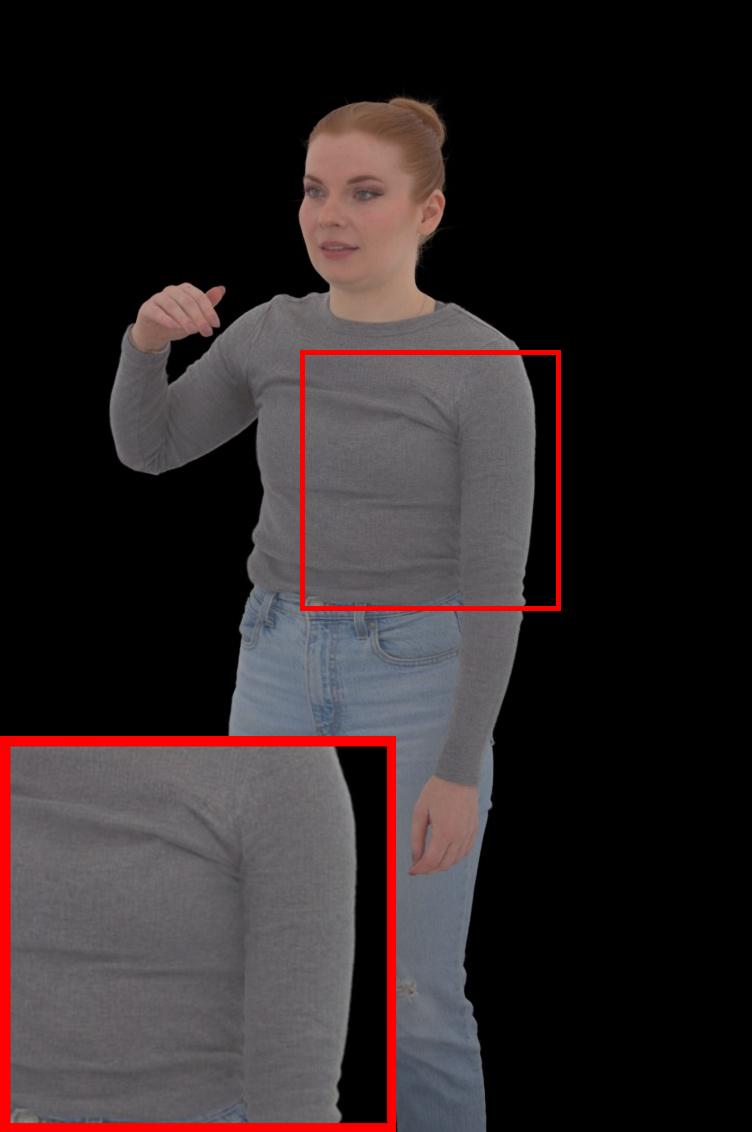} \\
        \includegraphics[width=0.24\textwidth]{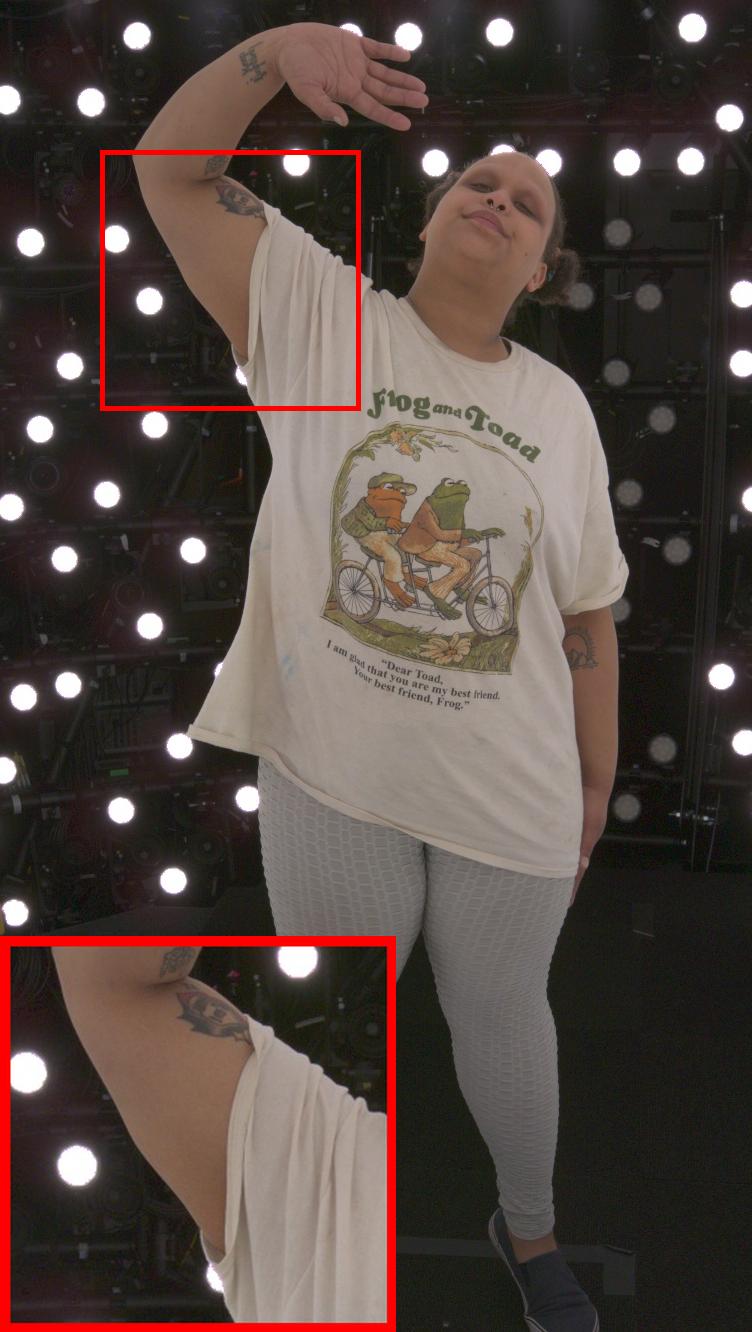} &
        \includegraphics[width=0.24\textwidth]{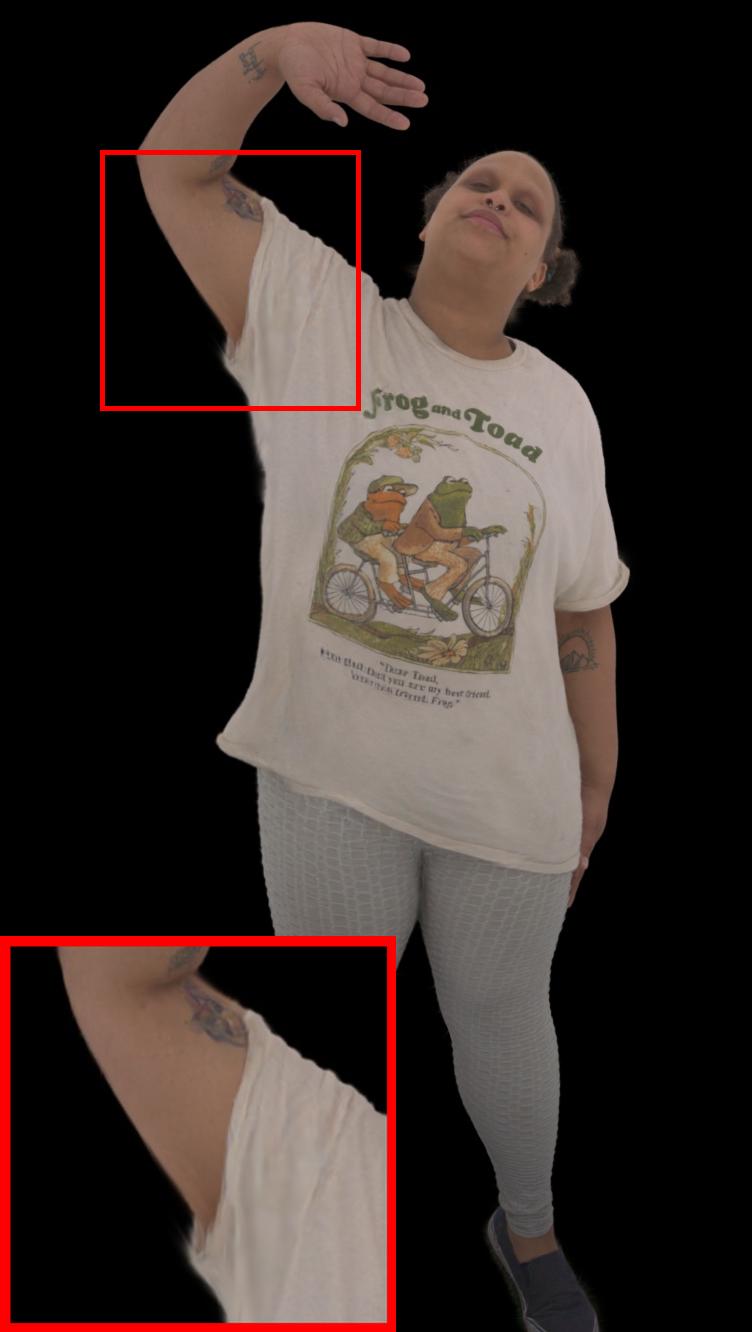} &
        \includegraphics[width=0.24\textwidth]{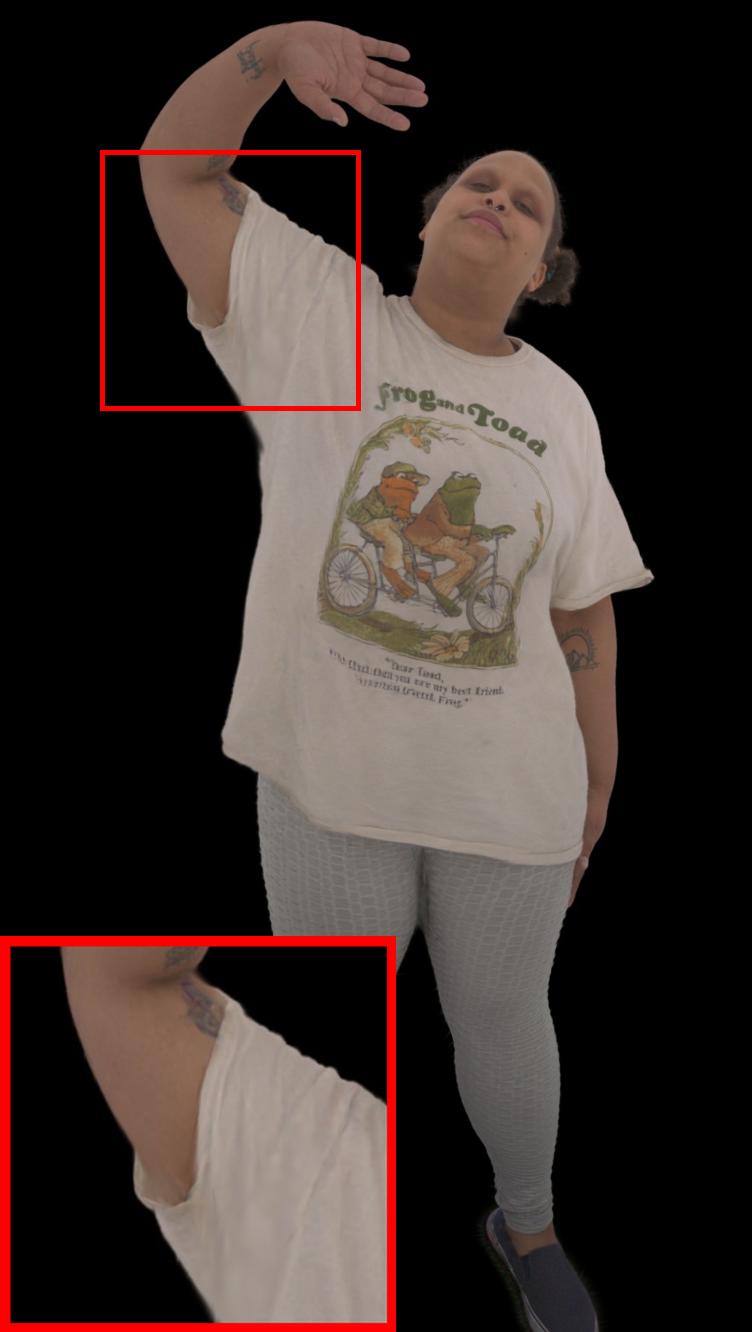} &
        \includegraphics[width=0.24\textwidth]{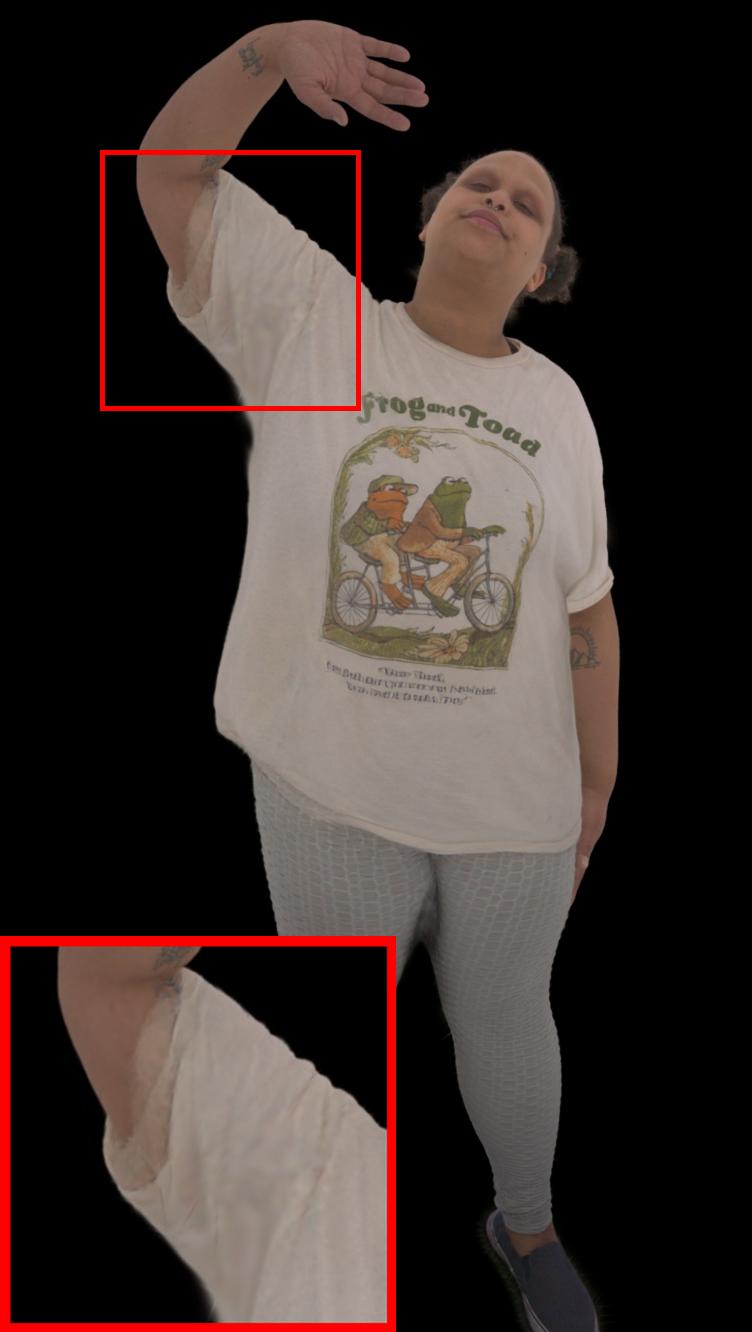} \\
        \includegraphics[width=0.24\textwidth]{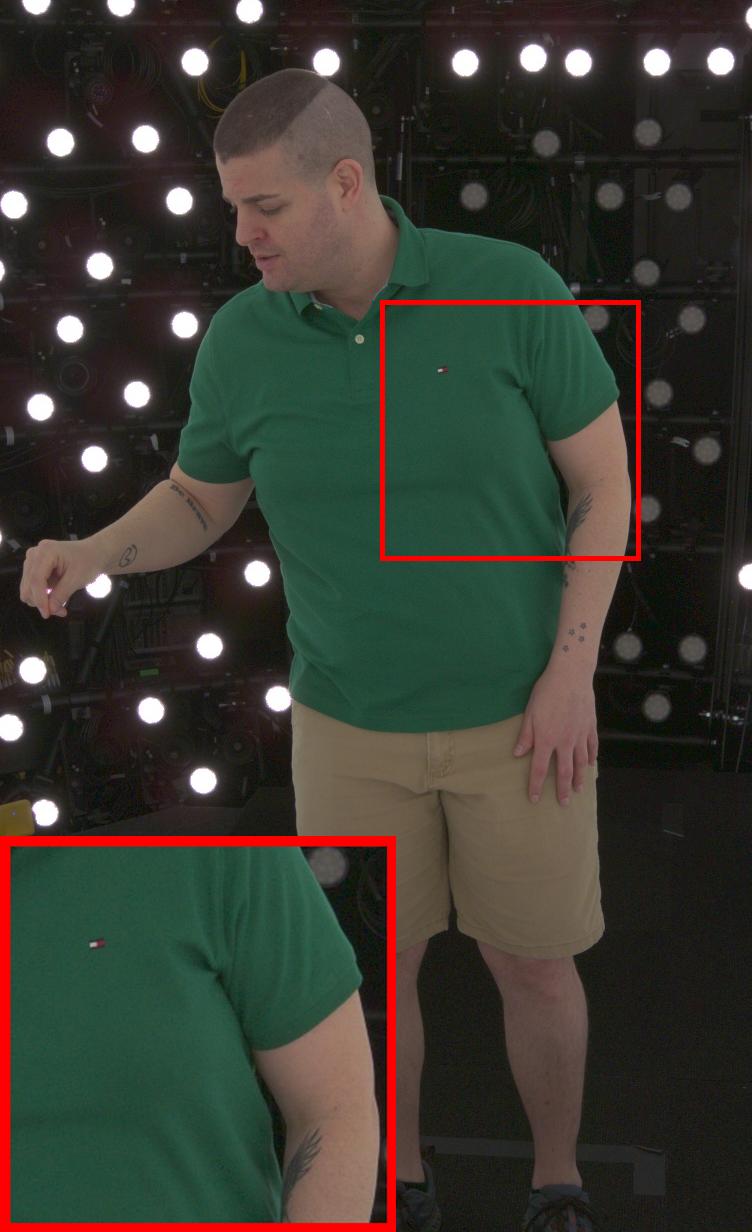} &
        \includegraphics[width=0.24\textwidth]{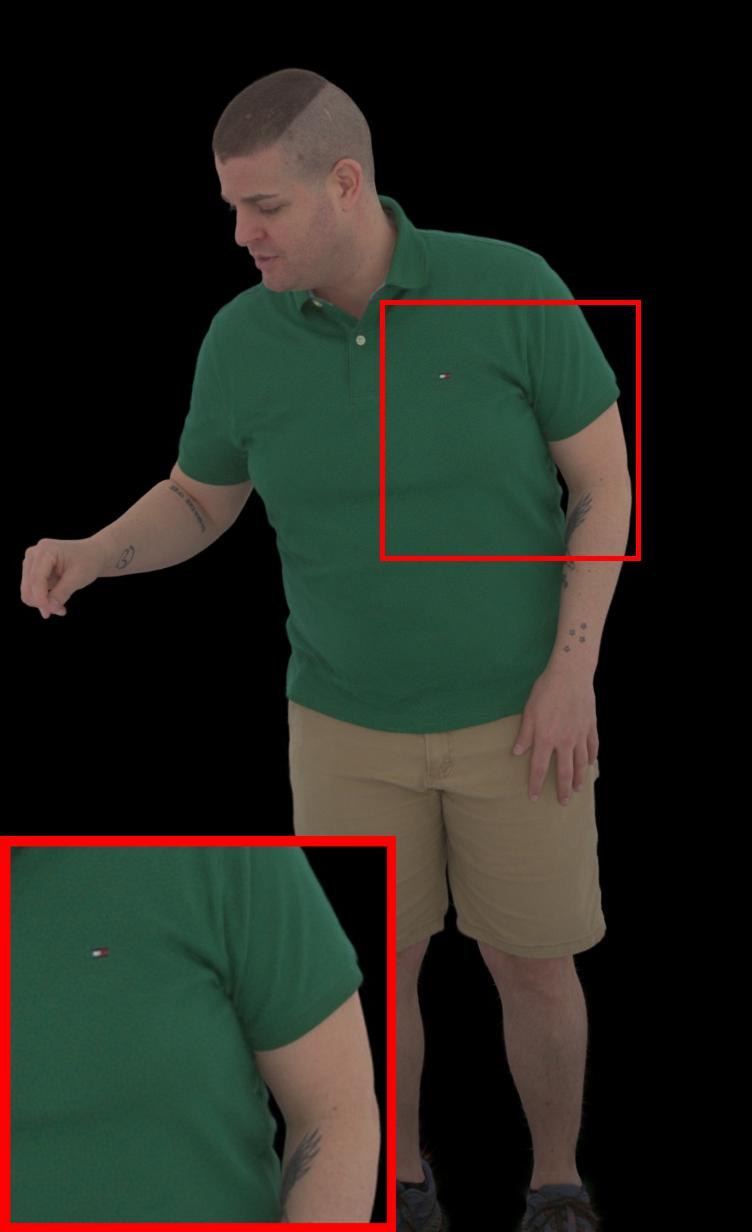} &
        \includegraphics[width=0.24\textwidth]{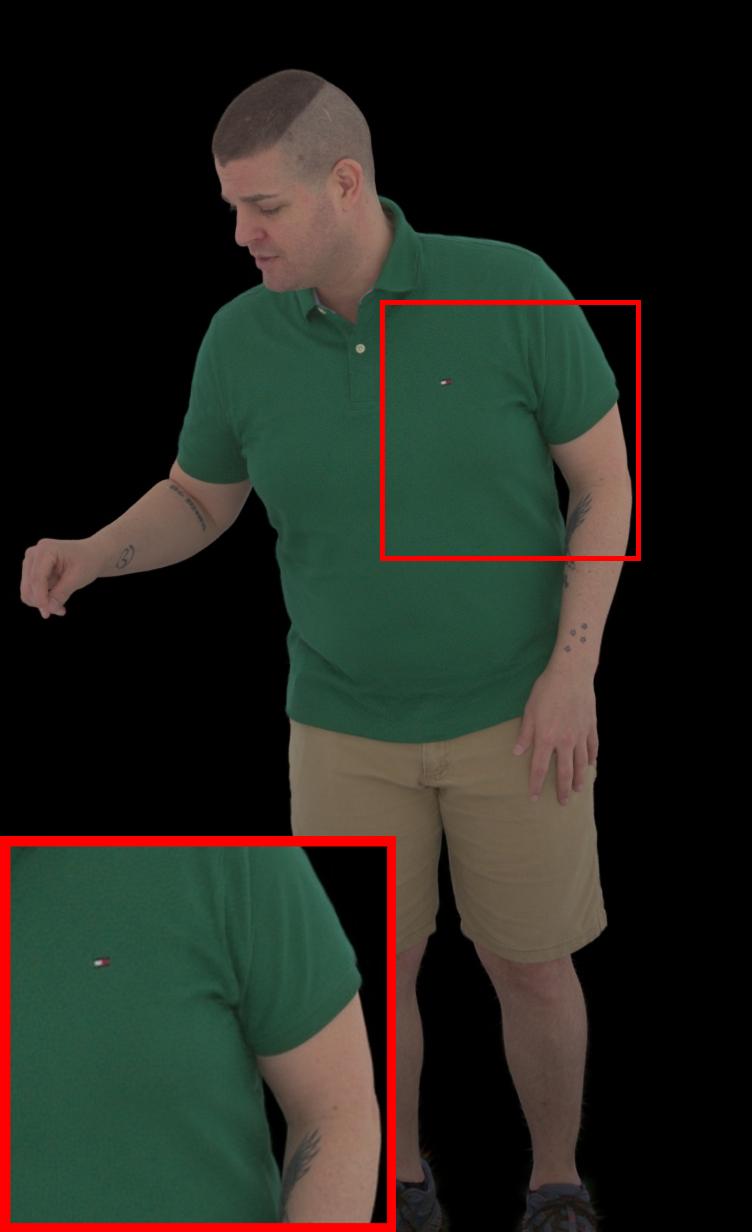} &
        \includegraphics[width=0.24\textwidth]{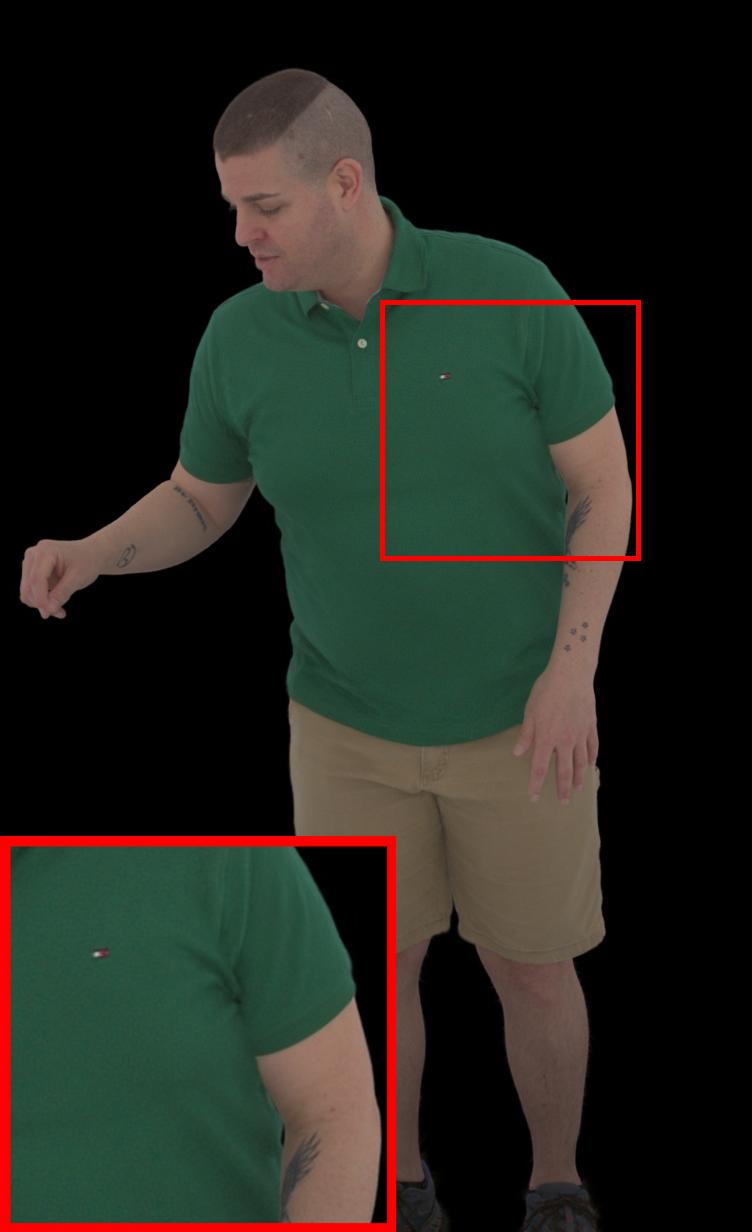} \\
        Ground truth & Ours (Encoder) & Ours (Predictor) & MMLPs\textsuperscript{$\dagger$}
    \end{tabular}
    \caption{Test-sequence comparison of our method using the texture encoder (unseen texture input) versus our appearance predictor (no textures), alongside MMLPs$^\dagger$.
    The encoder provides an upper bound closest to the ground truth, while the predictor yields temporally smooth, plausible appearance evolution without texture inputs.}
    \label{fig:results_collage}
\end{figure}

Finally, \cref{fig:results_collage} highlights (i) the generalization of our appearance encoder and (ii) the behavior of the appearance predictor on unseen sequences.
When provided with projected UV textures, the encoder produces latents that yield renderings closest to the ground truth, despite never observing these exact textures during training.
The predictor generates temporally smooth and plausible appearance changes without texture input; while it does not always match the ground-truth appearance (due to the inherent one-to-many ambiguity), it avoids the erratic switching and shading artifacts commonly observed in \emph{MMLPs$^\dagger$}.


\section{Limitations \& Future Work}

While our appearance predictor yields stable and plausible results in most cases, several limitations remain.
First, despite explicitly introducing appearance latents, our current formulation does not fully disentangle pose-driven effects from appearance variation.
In practice, the predictor may absorb factors that are, in principle, deterministically explainable from pose.
Our efforts of enforcing a stricter separation in our experiments---similar to \cite{bagautdinov2021drivingsignal}---consistently reduced reconstruction quality and limited the model's ability to fit fine details.

Second, strictly localized pose conditioning cannot represent certain long-range interactions, such as cast shadows of the arms onto the torso.
As a result, these effects can be encoded in the appearance latents, which may reduce robustness under extreme out-of-distribution motion.
A promising direction is modeling long-range shading separately---\eg, via an explicit, low-frequency lighting or shadow component~\cite{wang2025relightable}---thereby reducing the burden on the appearance codes and improving stability when extrapolating beyond the training distribution.

Finally, the initialization of the appearance predictor is currently simplistic: we either use encoder-produced codes when a UV texture is available, or fall back to a fixed zero code otherwise.
Although the predictor is trained to recover from this setting over the first few frames, it introduces a short ``ramp-up'' period before reaching a coherent appearance.
While this can be hidden from the user, a more principled generative initialization---\eg, conditioned on the first pose(s)---could yield realistic initial appearances without requiring this warm-up.

Future work should therefore focus on stronger pose--appearance disentanglement, explicit modeling of long-range shading effects, and improved test-time initialization for appearance prediction.

\section{Conclusion}

We presented a 3D Gaussian Splatting avatar model that improves both reconstruction fidelity and test-time stability by decoupling appearance variation from pose.
By learning per-frame appearance latents from reconstructed UV textures, our method builds a compact and semantically meaningful appearance space that captures time-varying details such as cloth, wrinkles, and hair.
We further showed that an autoregressive predictor operating in this latent space enables temporally coherent and visually plausible appearance evolution during driving without texture inputs.
Finally, our principled localization of pose parameters reduces spurious pose--appearance correlations, improving robustness under novel motions and yielding more stable downstream animation.



\clearpage

%
%
\bibliographystyle{splncs04}
\bibliography{main}

\clearpage

\appendix

\begin{center}
  {\LARGE\bfseries Supplementary Material\par}
  \vspace{0.5em}
  {\large {Autoregressive Appearance Prediction for 3D Gaussian Avatars}}
\end{center}

\section{Spatial MLPs Preliminaries}

Zhan et al.~\cite{zhan2025spatialmlps} construct a hierarchical structure of points by sampling a number of anchors/control points/Gaussians (300/10k/200k) on the template mesh.
During initialization, they calculate weights $t_{i,j}$ for the $i$-th Gaussian at position $\vect{x}_i$ to the three closest anchors in their neighborhood $\mathcal{N}_3$ based on the reciprocal distance $d_{i,j}$ to each anchor at position $\vect{x}^a_j$:
\begin{align}
t_{i,j}=\frac{d_{i,j}}{\sum_{k\in \mathcal{N}_3} d_{i,k}}, \quad \text{with} \quad d_{i,j} = \frac{1}{\|\vect{x}_i - \vect{x}^a_j\|_2}.
\label{eq:interpol_weights}
\end{align}
Each anchor holds an MLP mapping pose parameters $\vect{\theta}$ to corrective weights $\vect{w}^a \in \mathbb{R}^B$. Every Gaussian then computes its corrective weights $\vect{w}_i$ as the weighted average $\vect{w}_i=\sum_{j\in \mathcal{N}_3} t_{i,j} \vect{w}_j^a$. The corrective weights of the control points $\vect{w}^c$ are computed using the same procedure.
Additionally, each Gaussian holds offset vectors $\vect{\delta\Lambda_k}\in\mathbb{R}^B$ and a bias $\lambda_k$ for each of the Gaussian properties' values $\{\vect{r},\vect{s},\vect{c},o\}$.
The actual values of each property $p_k$ (e.g. scale in x-dimension $\vect{s_x}$) of the $i$-th Gaussian in canonical space are then computed as
\begin{align}
p_k=h_k(\lambda_k + \delta\lambda_k), \quad \text{with} \quad \delta\lambda_k = \langle \vect{w}_i, \vect{\delta\Lambda_k} \rangle,
\end{align}
with $h_k$ being a property-specific activation function.
Equivalently, each control point stores bias terms to compute its positional offset $\vect{\delta x^c}$.
Each Gaussian then computes its positional offset $\vect{\delta x}$ from the weighted average of the three closest control points $\mathcal{N}^c_3$ (similar to \cref{eq:interpol_weights} and $\vect{w}_i$) plus a small per-Gaussian offset $\vect{\delta x}_0$.
Adding this offset to the initial position of the Gaussian on the template mesh gives its final position in canonical space.
Finally, the canonical Gaussians are then posed using linear blend skinning (LBS).

\section{Dataset Details}

\subsection{Capture Info}

\begin{figure}[!ht]
    \centering
    \begin{subfigure}{0.25\linewidth}
        \centering
        \includegraphics[width=1.\linewidth]{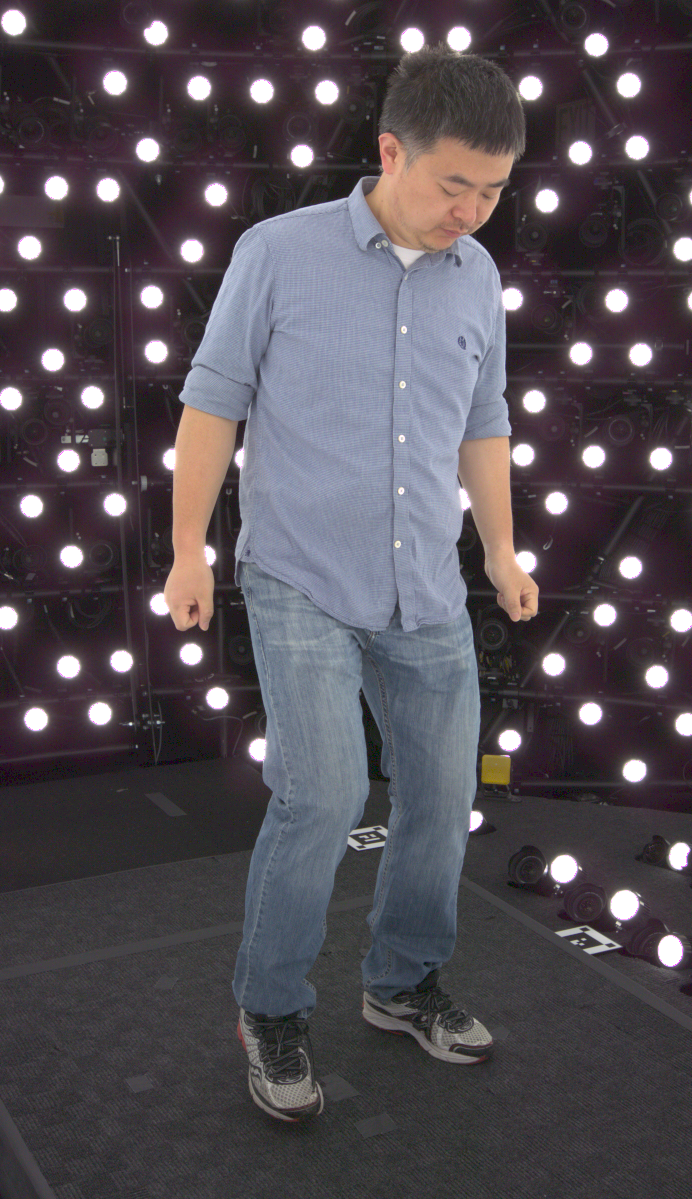}
        \caption{Actor 1}
    \end{subfigure}
    \begin{subfigure}{0.25\linewidth}
        \centering
        \includegraphics[width=1.\linewidth]{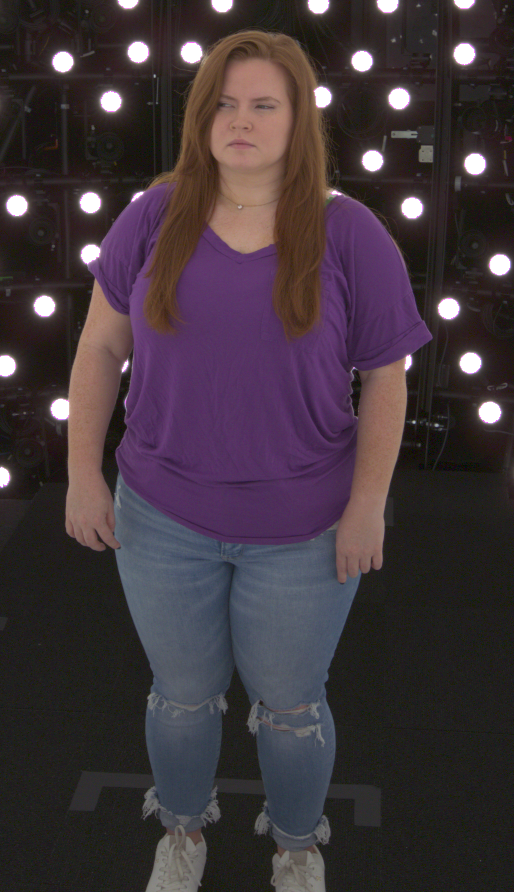}
        \caption{Actor 2}
    \end{subfigure}
    \begin{subfigure}{0.25\linewidth}
        \centering
        \includegraphics[width=1.\linewidth]{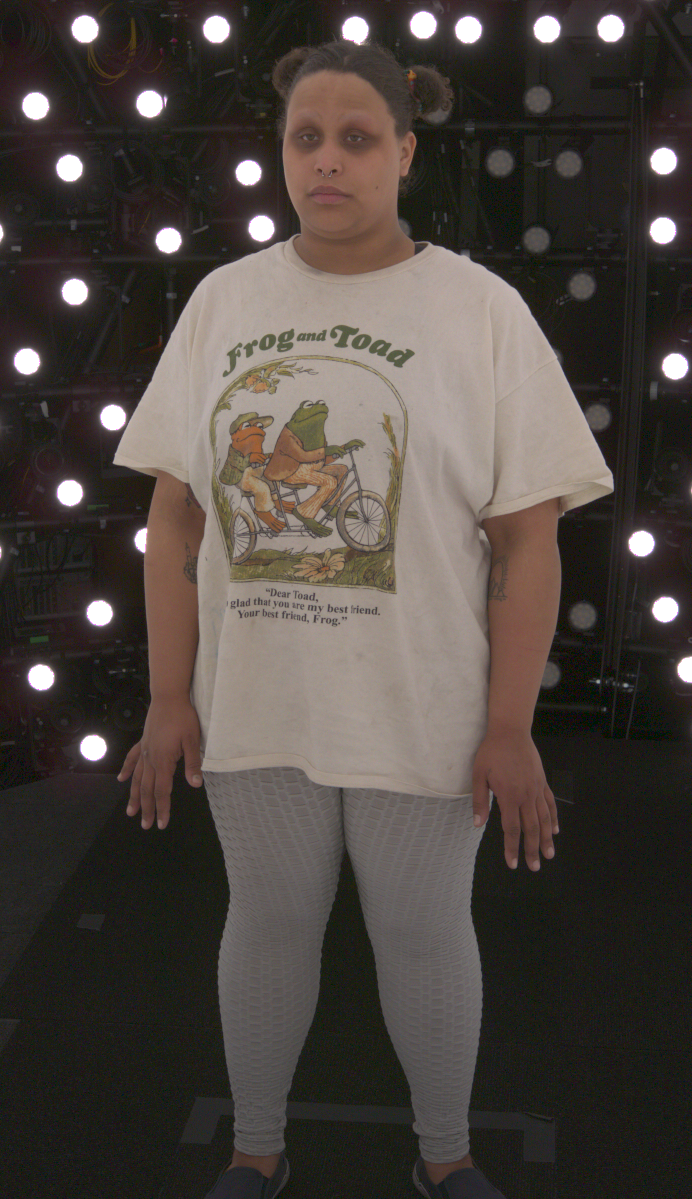}
        \caption{Actor 3}
    \end{subfigure}
    \begin{subfigure}{0.25\linewidth}
        \centering
        \includegraphics[width=1.\linewidth]{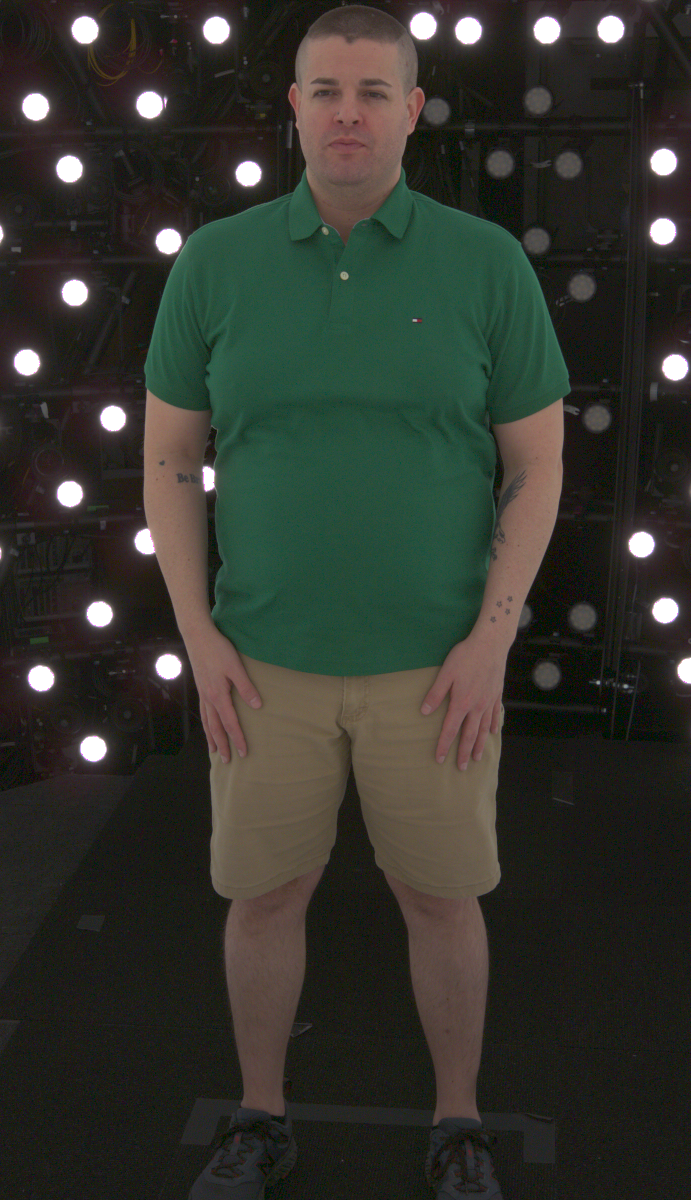}
        \caption{Actor 4}
    \end{subfigure}
    \begin{subfigure}{0.25\linewidth}
        \centering
        \includegraphics[width=1.\linewidth]{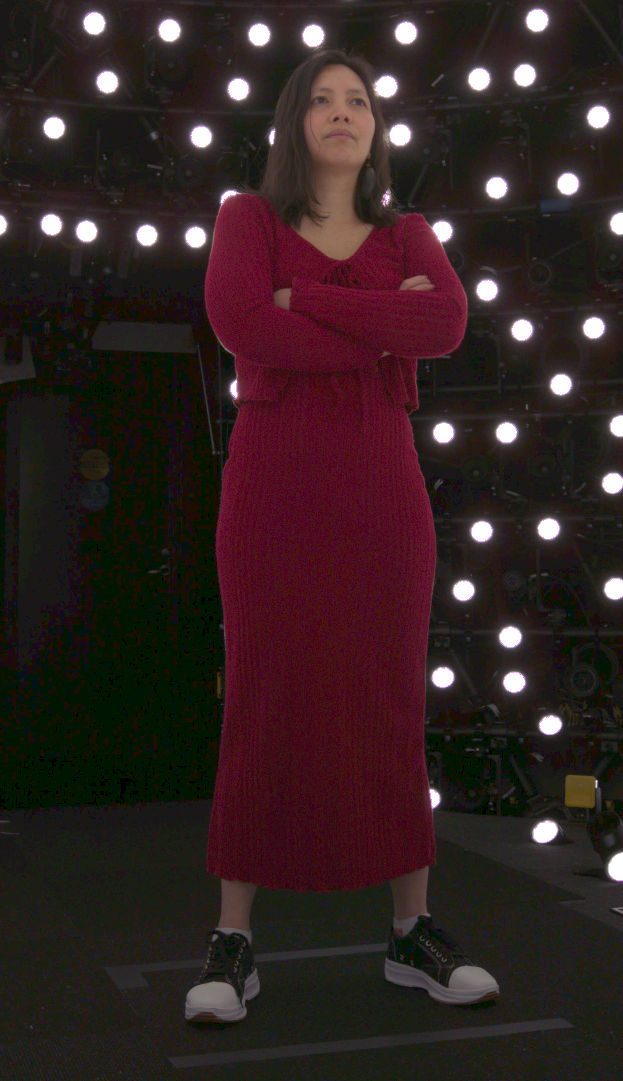}
        \caption{Actor 5}
    \end{subfigure}
    \begin{subfigure}{0.25\linewidth}
        \centering
        \includegraphics[width=1.\linewidth]{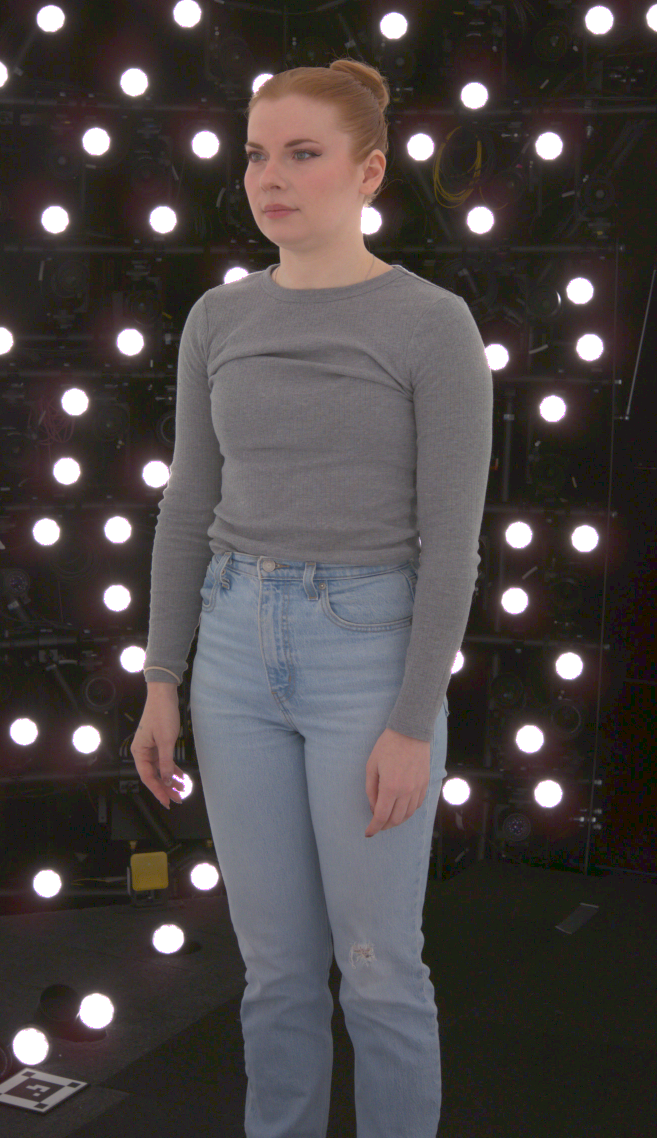}
        \caption{Actor 6}
    \end{subfigure}
    \caption{Our dataset consists of six captures of varying difficulty.}
    \label{fig:dataset_personas}
\end{figure}

\cref{fig:dataset_personas} presents the six captures evaluated in our experiments.
The captures span a range of difficulty factors, including long hair (\emph{Actor 2 and 5}), loose clothing (\emph{Actor 3 and 5}), untucked shirts (\emph{Actor 1 and 4}), and tight yet wrinkled garments (\emph{Actor 6}).
Each actor is recorded for ${\sim}$30 minutes, from which we select 12--20 minutes of segments exhibiting high pose diversity.
The captures further differ in frame rate, ranging from 24--45 FPS, which results in 20--35k training frames.
We evaluate the appearance predictor on the test sequences at the training frame rate.
Running the appearance predictor at a frame rate different from the one used during training may require additional measures.

\subsection{UV Texture Projection}

We use the mean shape of a subject as the template mesh and pose it using per-frame tracked body poses.
For each view, we rasterize the posed mesh and project the image pixels to the mesh's texels.
To minimize the impact of inaccurate surface boundaries, we apply eroded segmentation masks to identify non-boundary pixels and only project these pixels.
When fusing texel intensities from multiple views, we first compute the median and filter out values outside the range $[0.8, 1.1]$ times the median intensity. 
The final texel value is then calculated by averaging the remaining intensities. Through experimentation, we find that this approach achieves reasonable spatial and temporal consistency.

\section{Training Details}

\subsection{Gaussian Model}

Following Zhan et al.~\cite{zhan2025spatialmlps} we initialize 300 anchors, 10k control points, and 200 Gaussians on the template mesh.
Each anchor is associated with an MLP with hidden layer sizes $\{512,256,256,256\}$, which outputs 16 coefficients for the Gaussian corrective basis and 16 coefficients for the control point basis.
We apply nonlinearities to each Gaussian property: an exponential activation for scale, a sigmoid for opacity, and a tanh for spherical harmonics (scaled such that sh0 can represent RGB values in the range $[0,1]$).
We do not apply correctives to opacity, and we additionally max-clamp scale at $0.2$ to encourage a more coherent and stable representation.
Training is performed for 100k iterations; we increase the spherical-harmonics degree from 0 to 1 after 60k iterations and enable correctives after 500 iterations.

We use the following learning rates for the (bias term / corrective basis) of each property: scale ($5{\times}10^{-4}/1{\times}10^{-4}$), rotation ($5{\times}10^{-4}/1{\times}10^{-4}$), opacity ($5{\times}10^{-4}/-$), sh0 ($2.5{\times}10^{-3}/1{\times}10^{-4}$), shN ($2.5{\times}10^{-5}/2.5{\times}10^{-6}$), control point position ($1.6{\times}10^{-4},1.6{\times}10^{-5}$), and Gaussian position ($-/1{\times}10^{-4}$).
The anchor MLP is trained with a learning rate of $5{\times}10^{-4}$ and a weight decay of $1{\times}10^{-3}$.

\subsection{Appearance Encoder}

The encoder takes a $1024{\times}1024{\times}3$ RGB image as input and produces a $32{\times}32{\times}N_l$ feature map, where the size of the latent codes $N_l=16$.
We use a CNN composed of multiple down-convolution layers with kernel size $k=3$ and stride $2$, followed by 8-wide group normalization and ReLU activations.
The feature dimension is doubled at each layer (except for the first layer, which increases the dimension from $3$ to $8$), until reaching a maximum of $128$ channels and a spatial resolution of $32{\times}32$.
After downsampling, two separate $3{\times}3$ convolutional layers predict $N_l$-dimensional maps for $\mu$ and $\log(\sigma^2)$, from which we sample per-texel spatial latent codes from a normal distribution using the re-parametrization trick~\cite{kingma2013auto}.
The encoder is trained with a learning rate of $5{\times}10^{-4}$ and weight decay $1{\times}10^{-3}$.

\subsection{Appearance Predictor}

The appearance predictor transformer uses a single transformer block with 4 attention heads, token (embedding) dimension 128, and feed-forward hidden dimension 256.
Each pose token is produced by a dedicated shallow MLP, instantiated separately for each pose parameter and modality, while each appearance token is produced by a dedicated shallow MLP per anchor.
We apply sinusoidal absolute positional encoding in the transformer block.
The model output is generated by a linear projection head that maps the 128-dimensional token representations back to 16-dimensional per-anchor latent codes.
All components of the appearance predictor are trained using a learning rate of $1{\times}10^{-3}$.

\section{Additional Qualitative Evaluation}

\begin{figure}[!h]
    \centering
    \setlength{\tabcolsep}{0pt}
    \begin{tabular}{cccccccc}
        \includegraphics[width=0.16\linewidth]{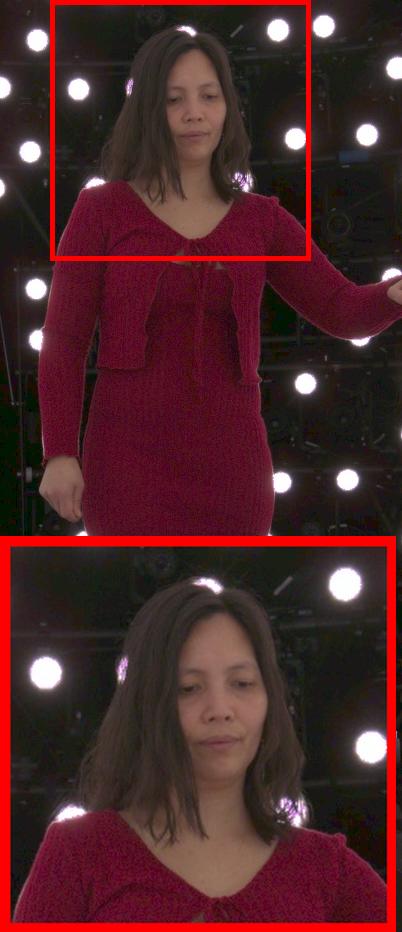} &
        \includegraphics[width=0.16\linewidth]{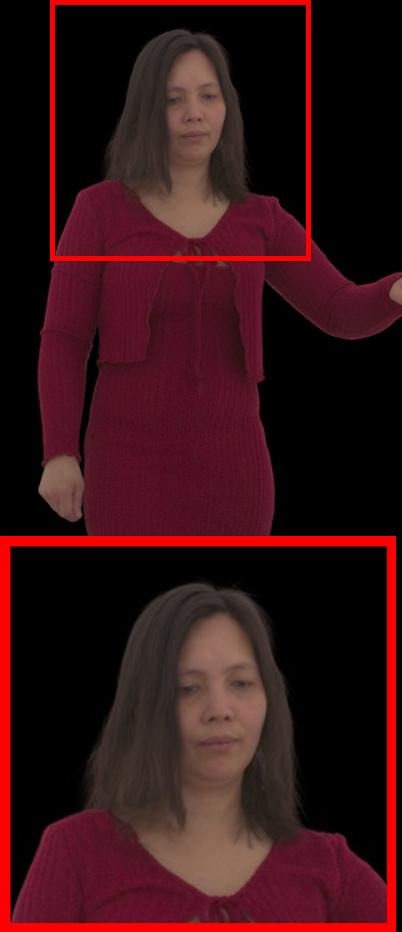} & \phantom{x}&
        \includegraphics[width=0.16\linewidth]{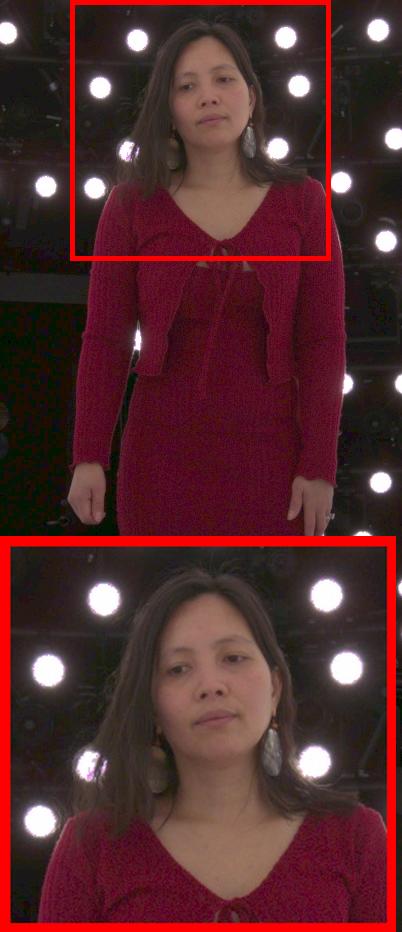} &
        \includegraphics[width=0.16\linewidth]{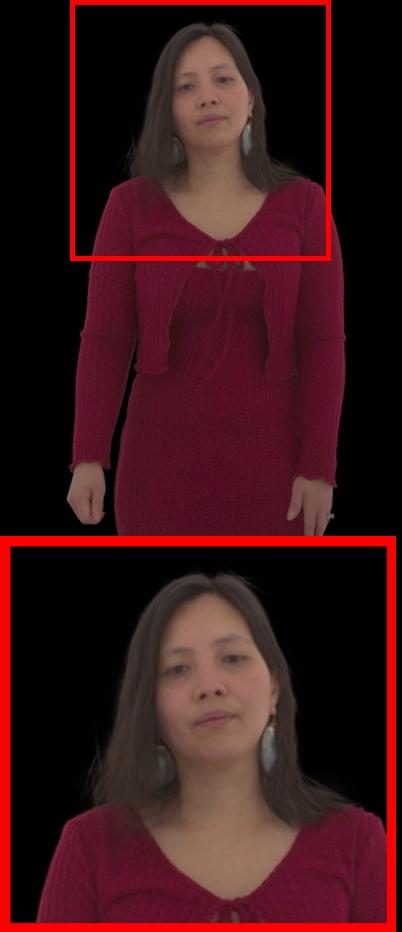} & \phantom{x} &
        \includegraphics[width=0.16\linewidth]{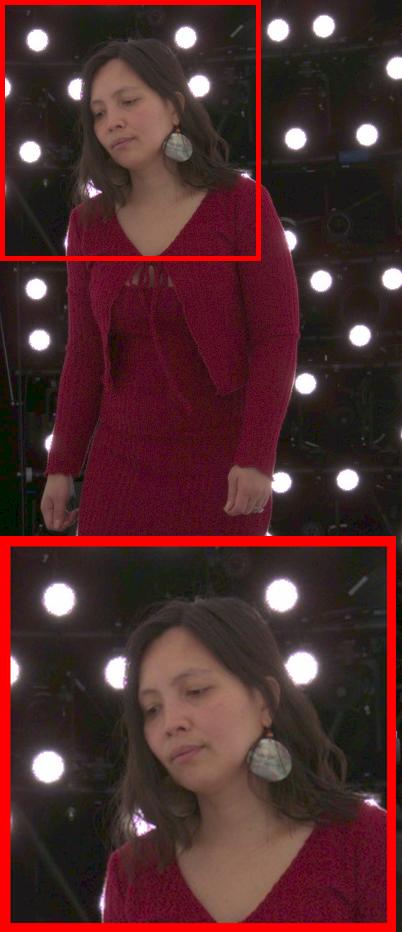} &
        \includegraphics[width=0.16\linewidth]{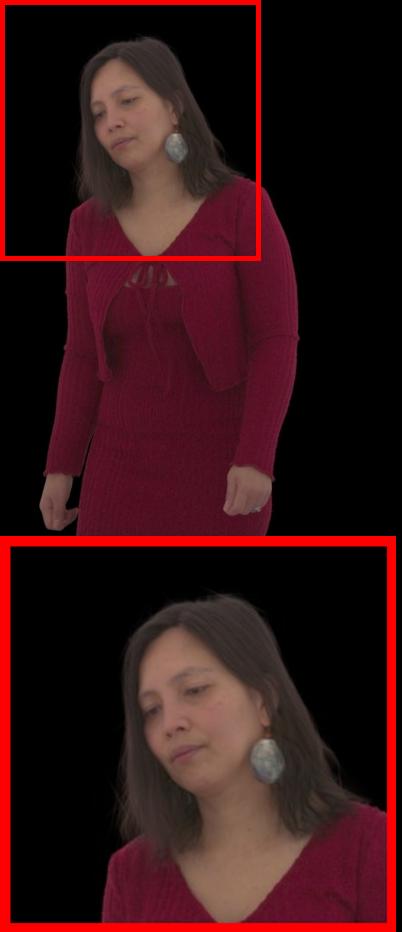} \\
        
        \includegraphics[width=0.16\linewidth]{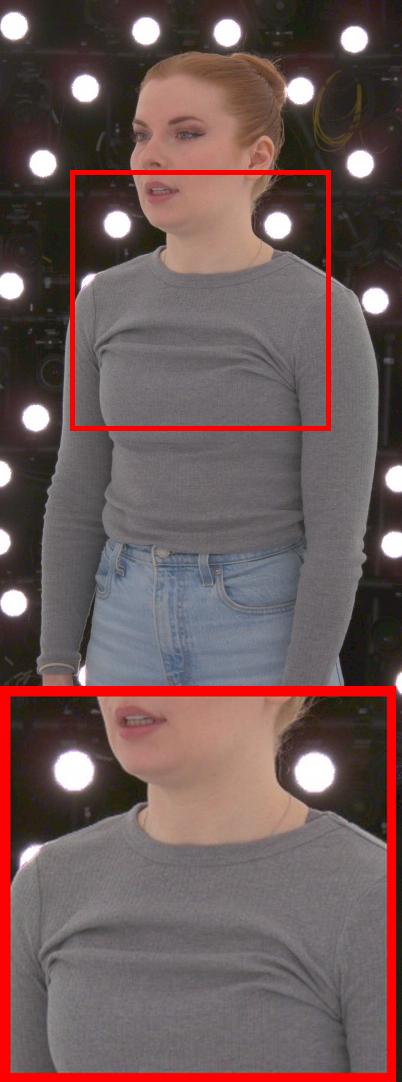} &
        \includegraphics[width=0.16\linewidth]{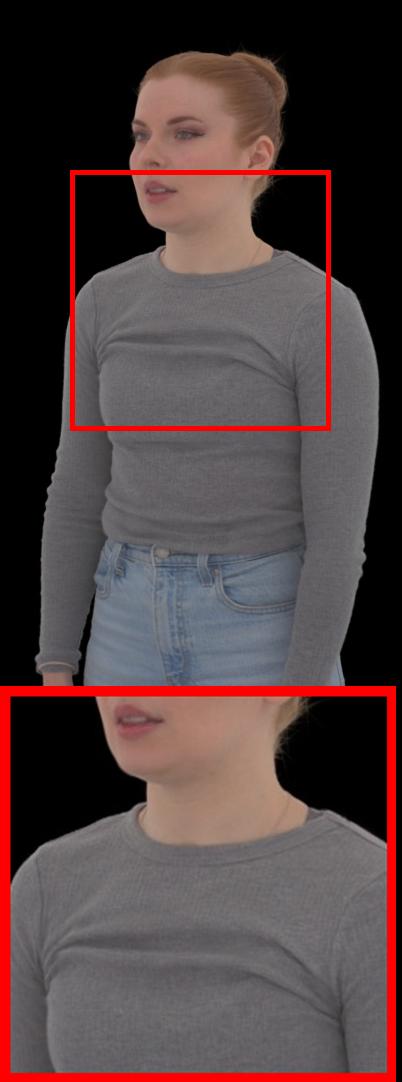} & &
        \includegraphics[width=0.16\linewidth]{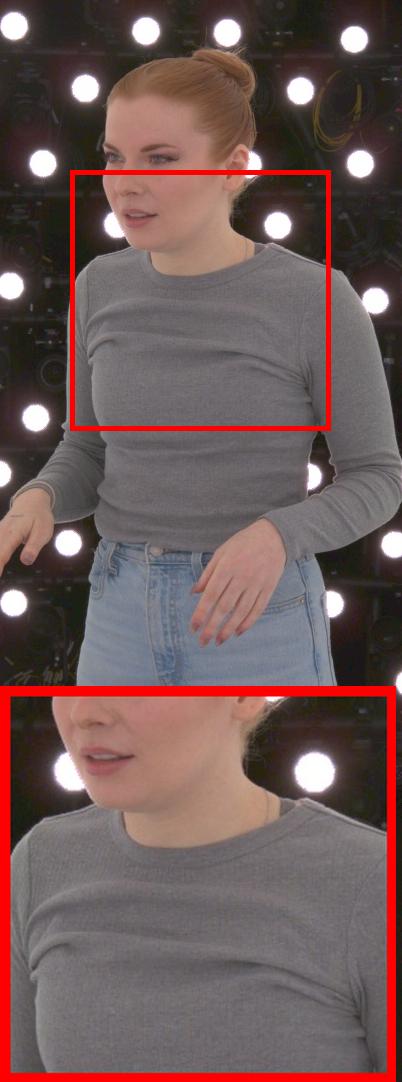} &
        \includegraphics[width=0.16\linewidth]{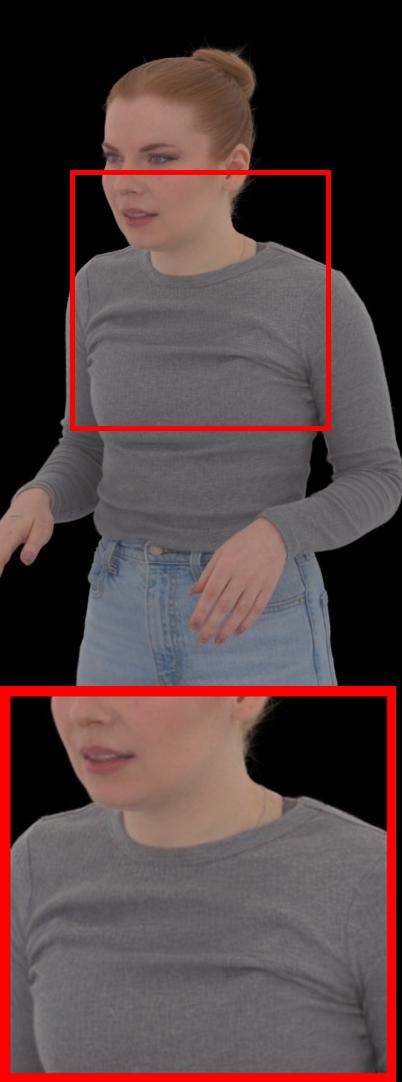} & &
        \includegraphics[width=0.16\linewidth]{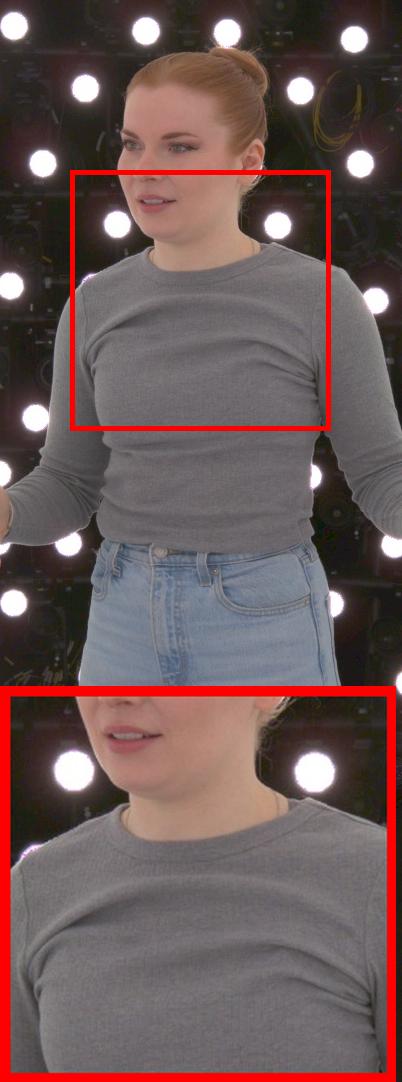} &
        \includegraphics[width=0.16\linewidth]{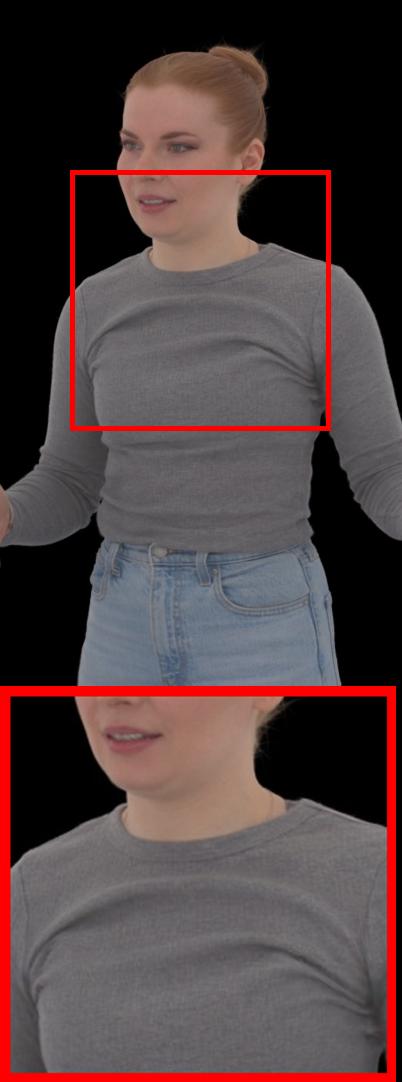}
    \end{tabular}
    \caption{
    Multiple image pairs showing the generalization capabilities of our texture encoder on the test set, \ie, novel poses and unseen UV textures. The left image shows ground truth and the right image is the appearance encoder result.
    }
    \label{fig:generalizability_collage}
\end{figure}

We evaluate the generalization capabilities of our encoder to unseen textures in \cref{fig:generalizability_collage}.
Our encoder is able to deliver results that are close to the ground truth, even though it might not have seen the exact hair configuration or wrinkle pattern during training.

\begin{figure}[!h]
    \centering
    \begin{tabular}{cccc}
        \includegraphics[trim=1cm 2cm 1cm 2cm,clip,width=0.24\textwidth]{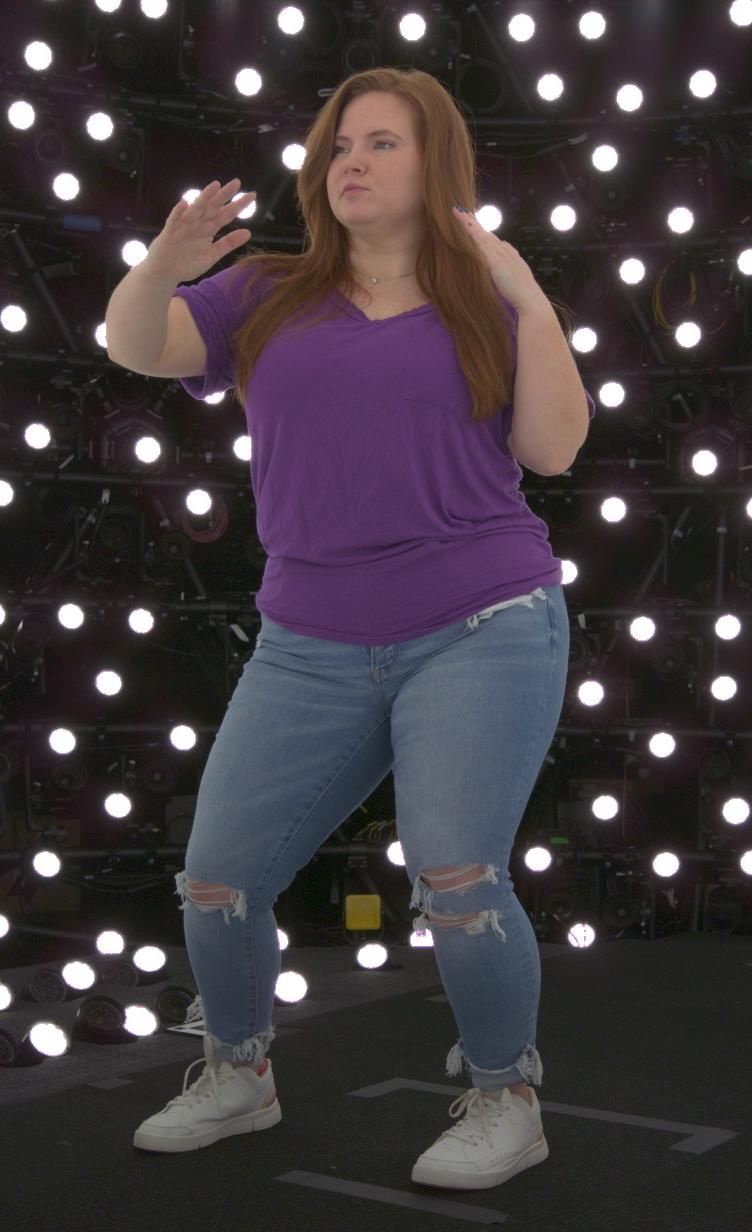} &
        \includegraphics[trim=1cm 2cm 1cm 2cm,clip,width=0.24\textwidth]{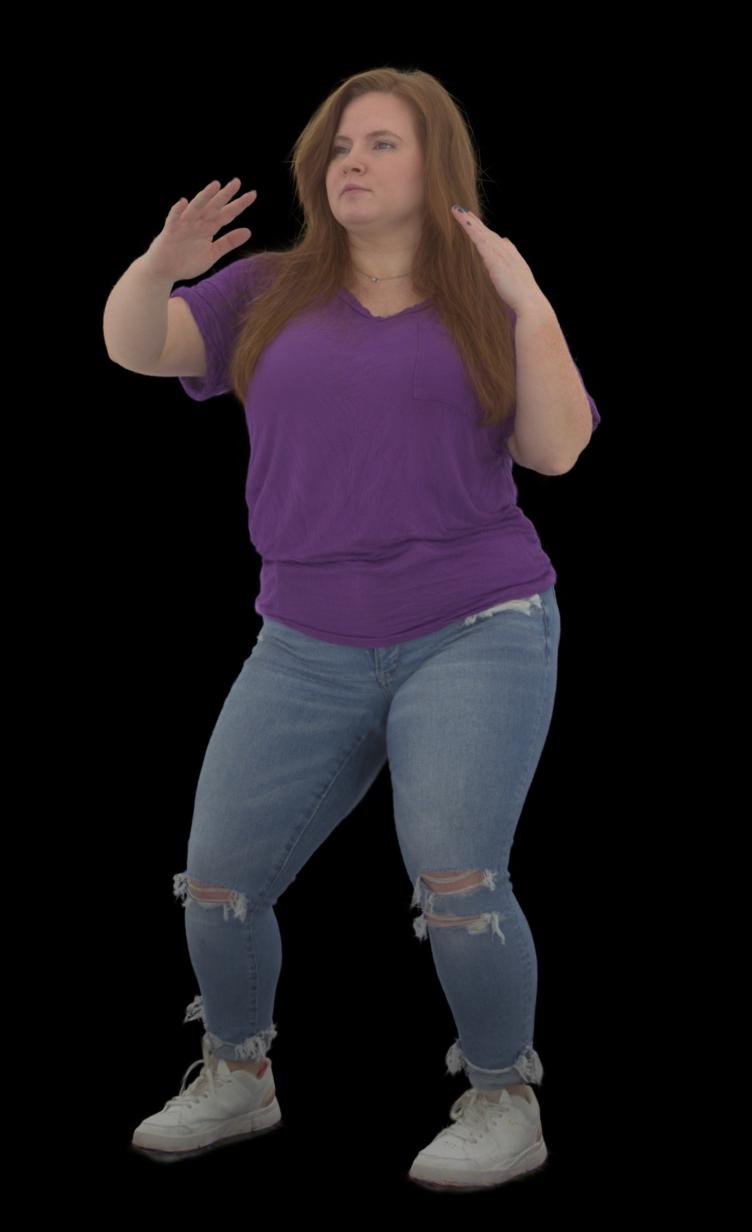} &
        \includegraphics[trim=1cm 2cm 1cm 2cm,clip,width=0.24\textwidth]{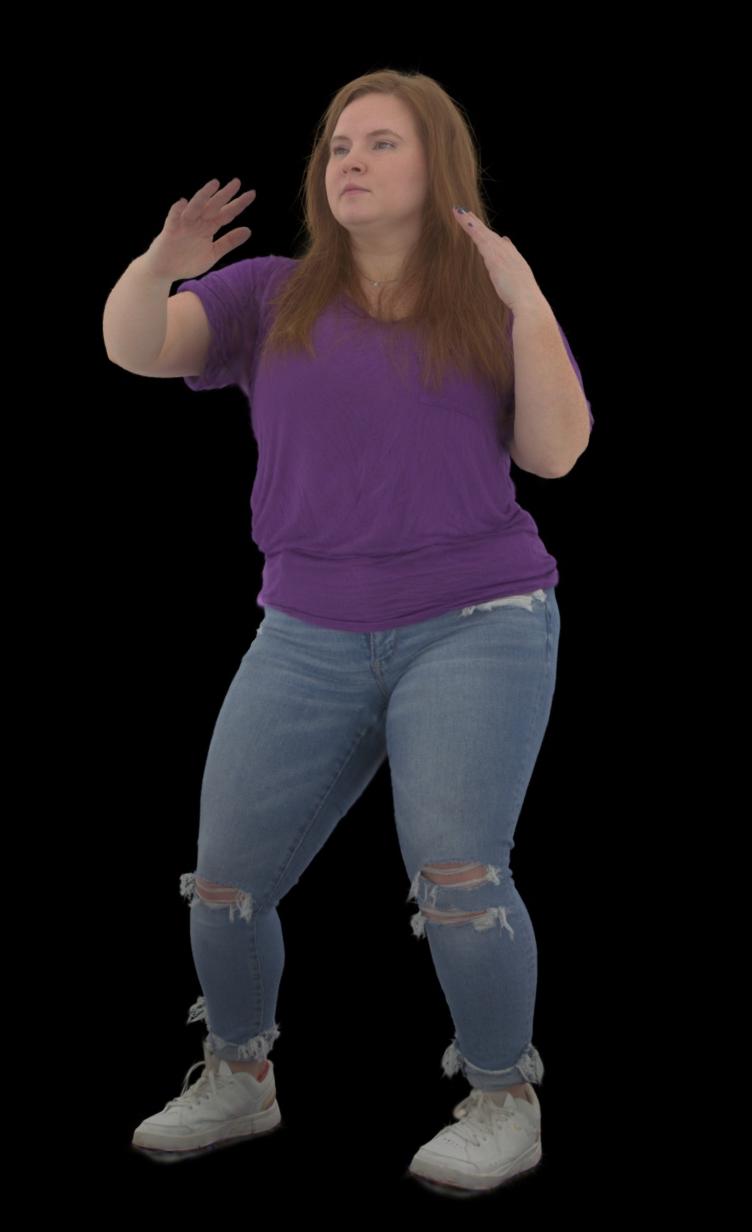} &
        \includegraphics[trim=1cm 2cm 1cm 2cm,clip,width=0.24\textwidth]{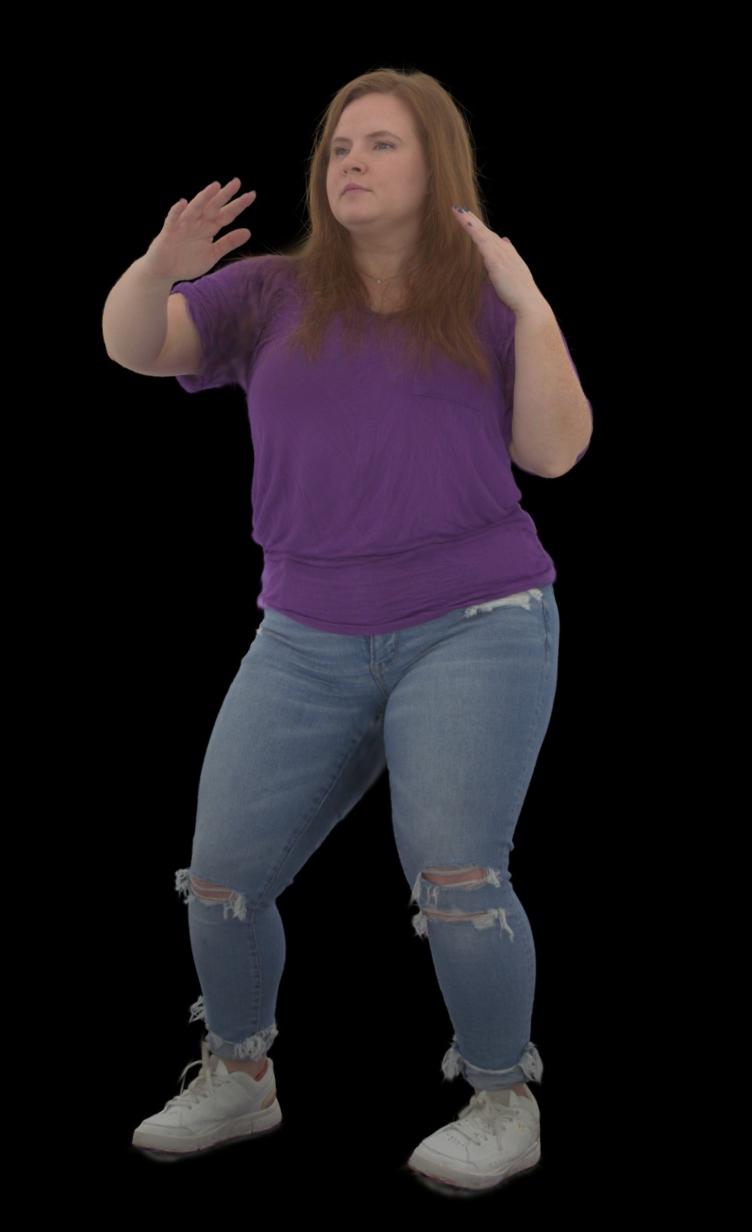} \\
        \includegraphics[trim=1cm 3cm 2cm 2cm,clip,width=0.24\textwidth]{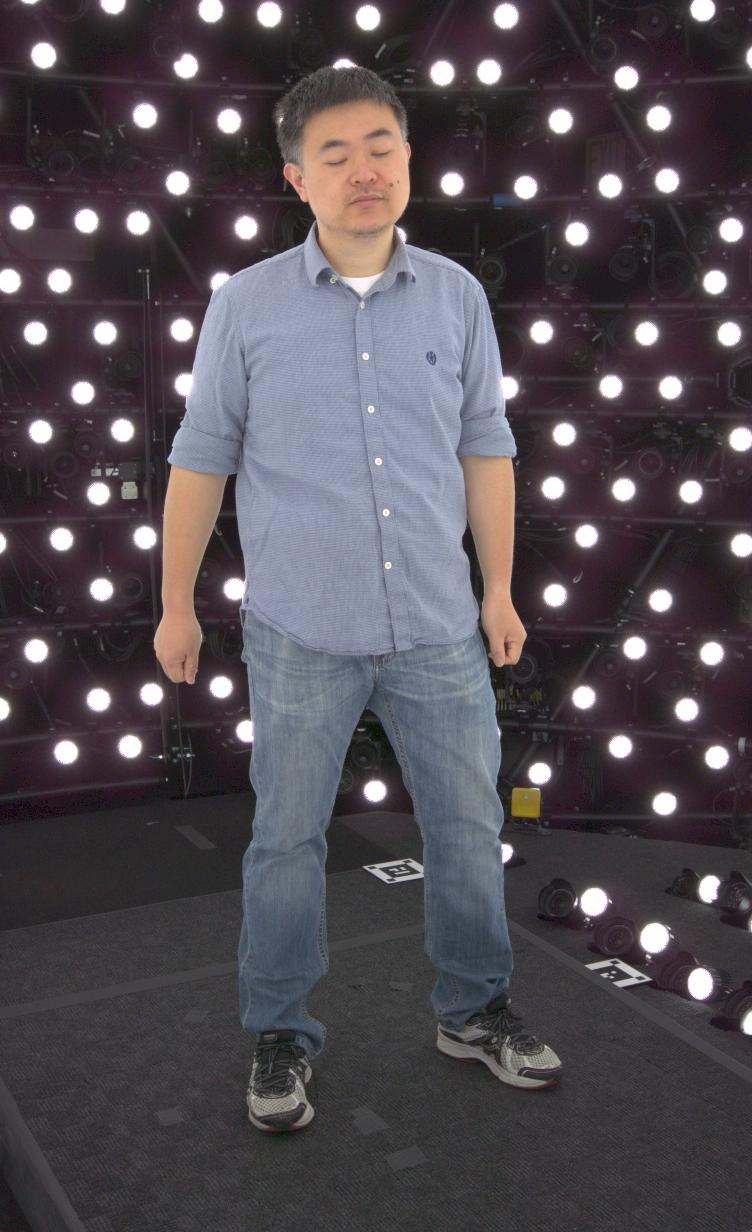} &
        \includegraphics[trim=1cm 3cm 2cm 2cm,clip,width=0.24\textwidth]{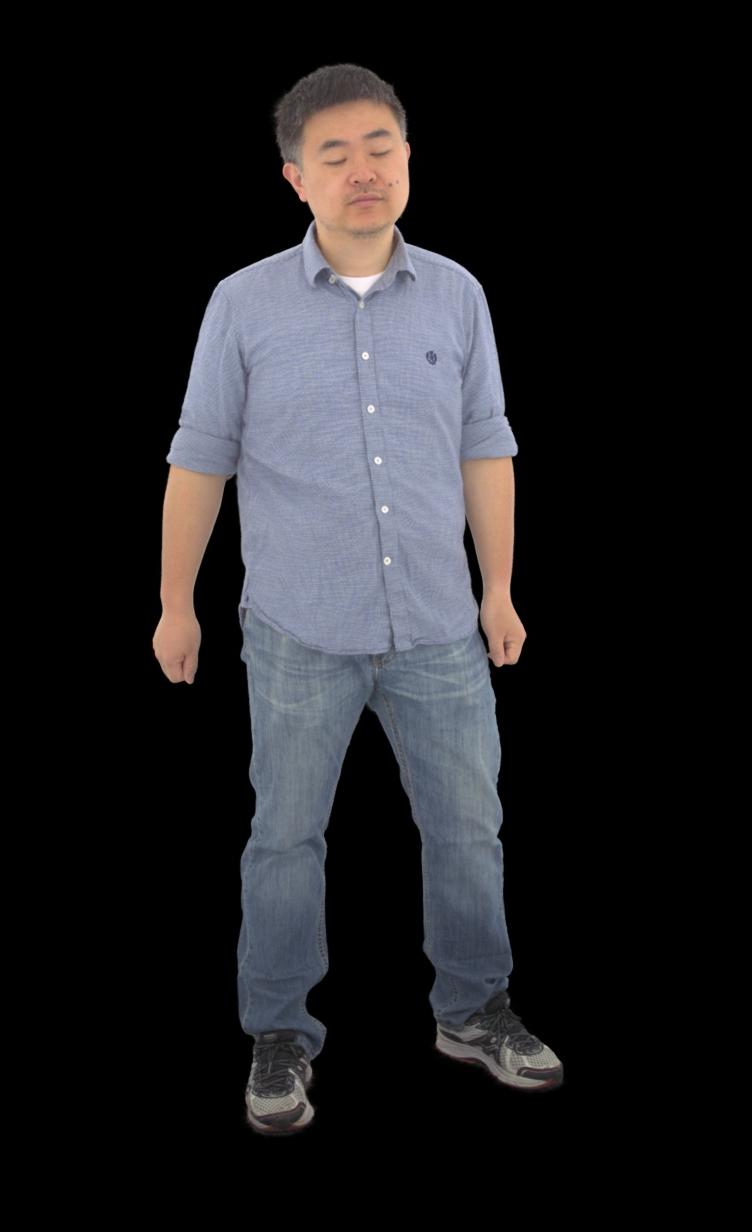} &
        \includegraphics[trim=1cm 3cm 2cm 2cm,clip,width=0.24\textwidth]{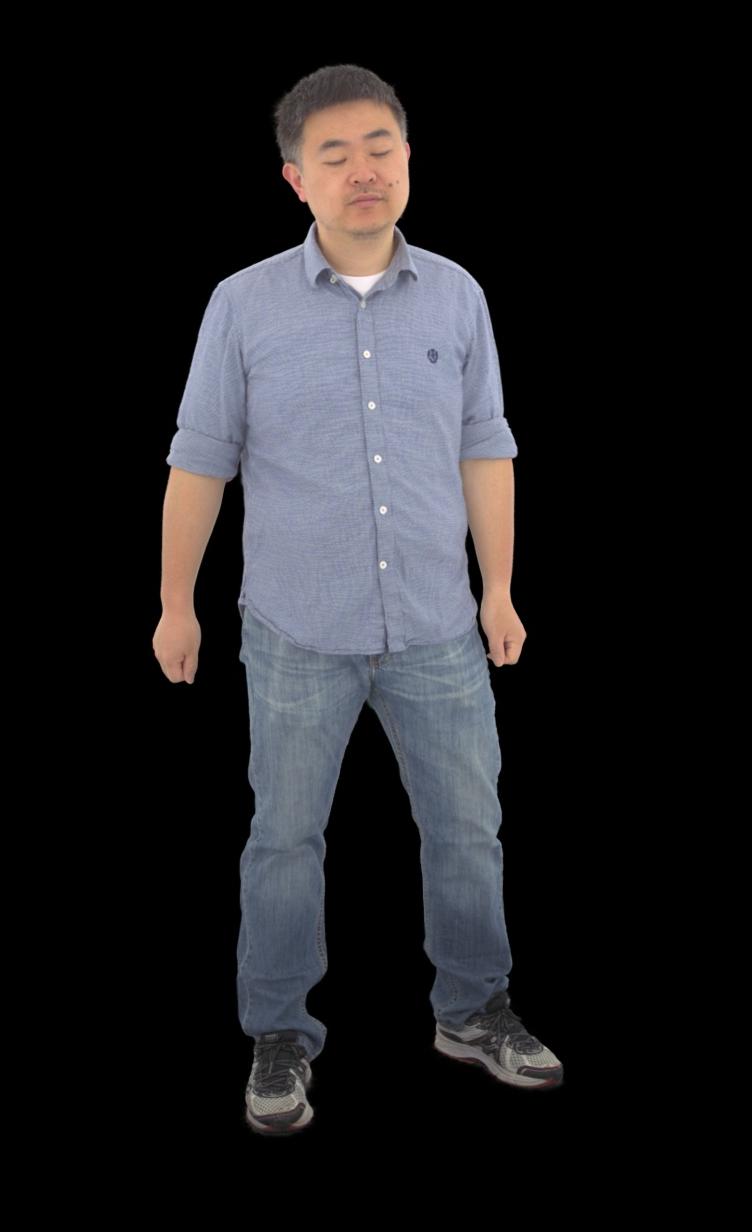} &
        \includegraphics[trim=1cm 3cm 2cm 2cm,clip,width=0.24\textwidth]{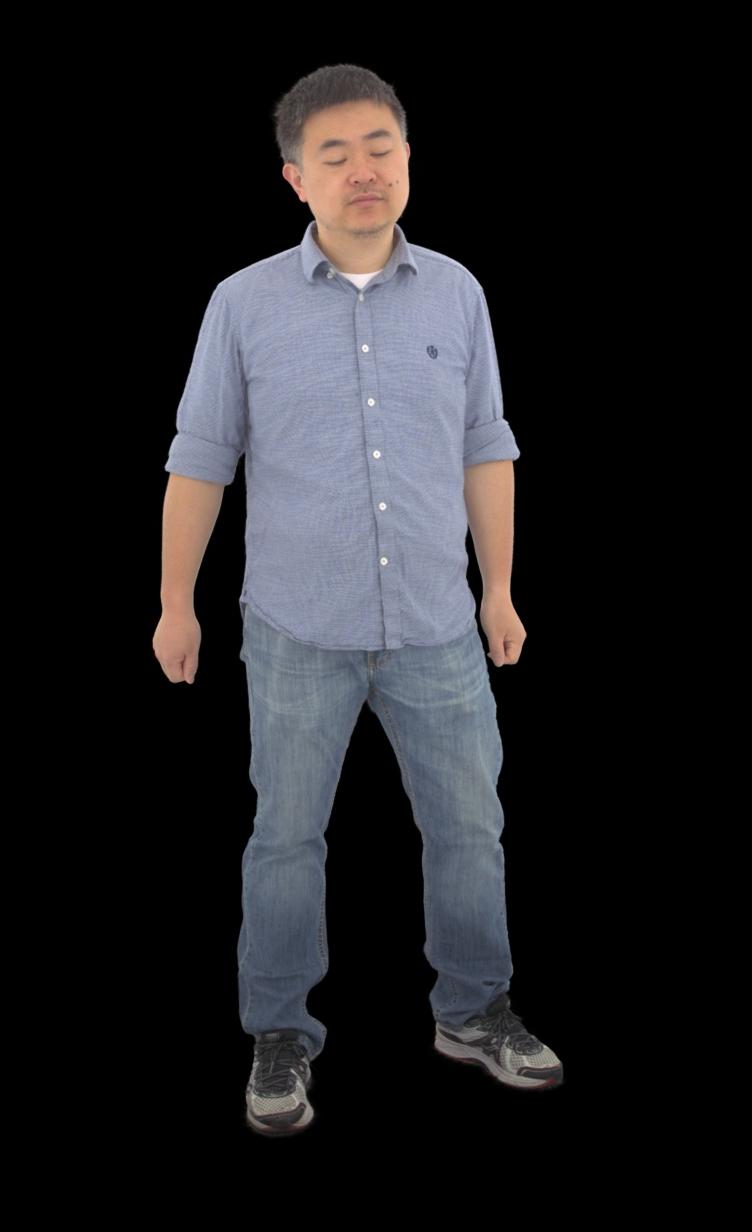} \\
        \includegraphics[width=0.24\textwidth]{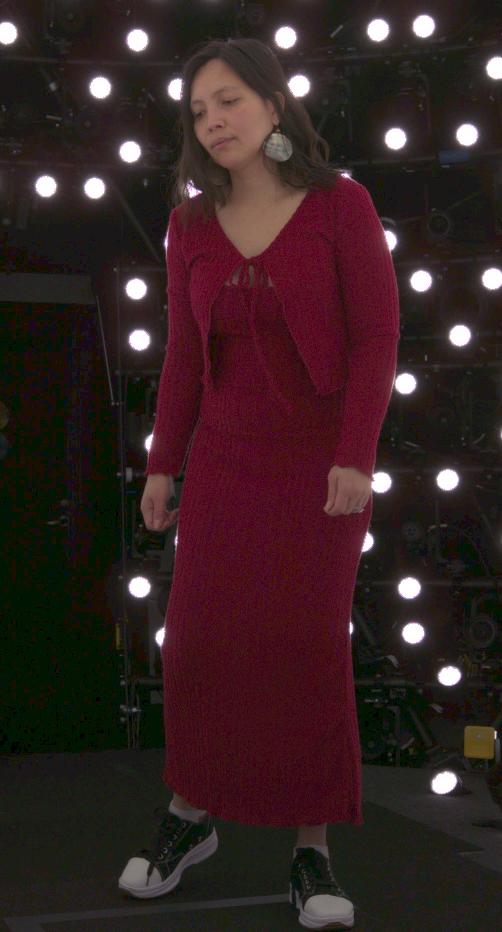} &
        \includegraphics[width=0.24\textwidth]{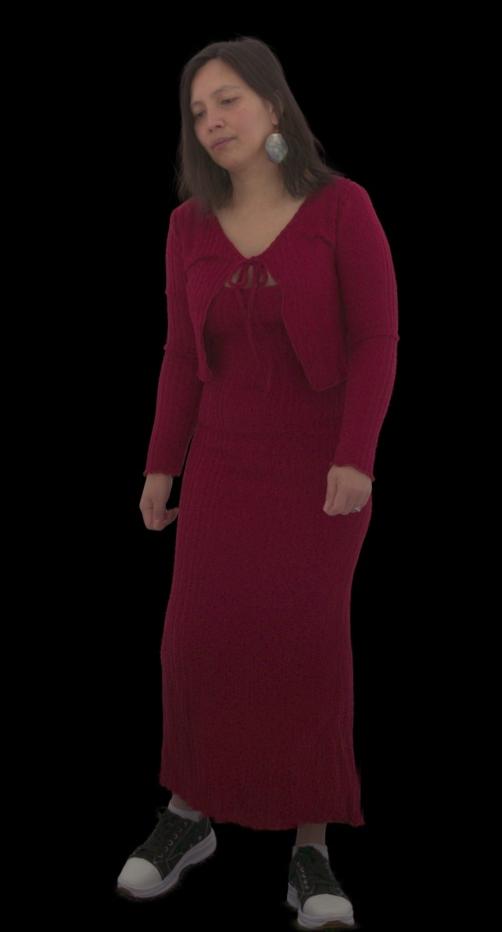} &
        \includegraphics[width=0.24\textwidth]{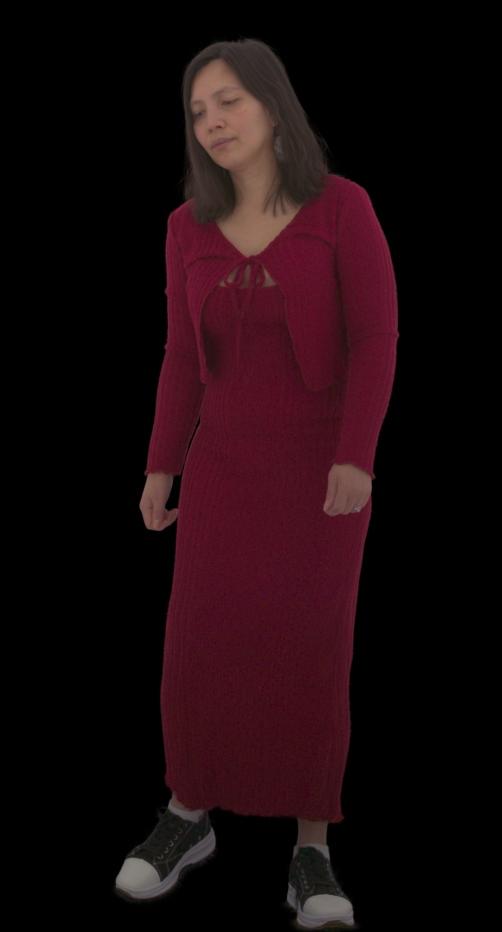} &
        \includegraphics[width=0.24\textwidth]{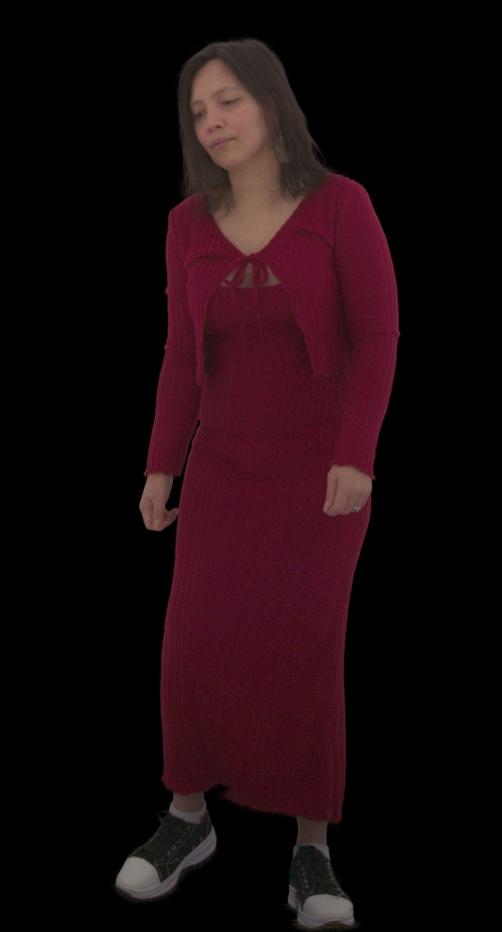} \\
        Ground truth & Ours (Encoder) & Ours (Predictor) & \emph{MMLPs$^\dagger$}
    \end{tabular}
    \caption{Test-sequence comparison of our method using the texture encoder (unseen texture input) versus our appearance predictor (no textures), alongside MMLPs$^\dagger$.}
    \label{fig:results_collage_other}
\end{figure}

We show additional results in \cref{fig:results_collage_other} for our encoder and the appearance predictor on a test sequence, compared to the baseline \emph{MMLPs$^\dagger$}~\cite{zhan2025spatialmlps}.
While the results from our appearance encoder are closer to the ground truth, the predictor results provide a believable appearance evolution and are more temporally stable than \emph{MMLPs$^\dagger$}, which suffers from spurious correlations and pose--appearance ambiguities.

\section{Per-Capture Quantitative Results}

We show the per-capture quantitative results on the test and training set against \emph{MMLPs$^\dagger$}~\cite{zhan2025spatialmlps} and \emph{nRFGCA$^\dagger$}~\cite{wang2025relightable} in \cref{tab:quant_percapture_test_pred} and \cref{tab:quant_percapture_train}.

\begin{table}[!h]
    \centering
    \scriptsize
    \caption{Test set image metrics on all compared methods. We initialize the appearance predictor for Ours with the encoder output and show additional results for re-initializing every 30 frames, and for using the texture encoder on the test sequences.}
    \setlength{\tabcolsep}{3pt}
\begin{tabular}{lcccccc|c}
\textbf{PSNR$^\uparrow$} & Actor 1 & Actor 2 & Actor 3 & Actor 4 & Actor 5 & Actor 6 & Mean \\ \hline
Ours & \cellcolor{best!40}31.52 & \cellcolor{best!15}31.90 & \cellcolor{best!15}28.82 & \cellcolor{best!40}35.03 & \cellcolor{best!40}34.38 & \cellcolor{best!40}36.31 & \cellcolor{best!40}32.99 \\
MMLPs$^\dagger$ & \cellcolor{best!00}31.29 & \cellcolor{best!00}31.79 & \cellcolor{best!00}28.41 & \cellcolor{best!00}33.68 & \cellcolor{best!15}34.31 & \cellcolor{best!00}34.55 & \cellcolor{best!00}32.34 \\
nRFGCA$^\dagger$ & \cellcolor{best!15}31.47 & \cellcolor{best!40}31.94 & \cellcolor{best!40}28.95 & \cellcolor{best!15}34.84 & \cellcolor{best!00}34.29 & \cellcolor{best!15}34.97 & \cellcolor{best!15}32.74 \\ \hline
Ours (re-init. 30) & \cellcolor{best!00}32.39 & \cellcolor{best!00}32.93 & \cellcolor{best!00}30.04 & \cellcolor{best!00}35.80 & \cellcolor{best!00}35.02 & \cellcolor{best!00}36.92 & \cellcolor{best!00}33.85 \\
Ours (encoder) & \cellcolor{best!00}33.33 & \cellcolor{best!00}33.85 & \cellcolor{best!00}31.48 & \cellcolor{best!00}36.71 & \cellcolor{best!00}35.63 & \cellcolor{best!00}37.67 & \cellcolor{best!00}34.78 \\
\\
\textbf{SSIM$^\uparrow$} & Actor 1 & Actor 2 & Actor 3 & Actor 4 & Actor 5 & Actor 6 & Mean \\ \hline
Ours & \cellcolor{best!15}0.923 & \cellcolor{best!40}0.942 & \cellcolor{best!15}0.928 & \cellcolor{best!40}0.950 & \cellcolor{best!15}0.931 & \cellcolor{best!40}0.959 & \cellcolor{best!40}0.939 \\
MMLPs$^\dagger$ & \cellcolor{best!00}0.922 & \cellcolor{best!00}0.941 & \cellcolor{best!00}0.927 & \cellcolor{best!00}0.947 & \cellcolor{best!00}0.930 & \cellcolor{best!00}0.953 & \cellcolor{best!00}0.937 \\
nRFGCA$^\dagger$ & \cellcolor{best!40}0.924 & \cellcolor{best!40}0.942 & \cellcolor{best!40}0.929 & \cellcolor{best!40}0.950 & \cellcolor{best!40}0.932 & \cellcolor{best!15}0.956 & \cellcolor{best!40}0.939 \\ \hline
Ours (re-init. 30) & \cellcolor{best!00}0.927 & \cellcolor{best!00}0.946 & \cellcolor{best!00}0.934 & \cellcolor{best!00}0.952 & \cellcolor{best!00}0.932 & \cellcolor{best!00}0.962 & \cellcolor{best!00}0.942 \\
Ours (encoder) & \cellcolor{best!00}0.933 & \cellcolor{best!00}0.950 & \cellcolor{best!00}0.942 & \cellcolor{best!00}0.954 & \cellcolor{best!00}0.934 & \cellcolor{best!00}0.965 & \cellcolor{best!00}0.946 \\
\\
\textbf{LPIPS$^\downarrow$} & Actor 1 & Actor 2 & Actor 3 & Actor 4 & Actor 5 & Actor 6 & Mean \\ \hline
Ours & \cellcolor{best!40}0.062 & \cellcolor{best!40}0.064 & \cellcolor{best!40}0.082 & \cellcolor{best!40}0.066 & \cellcolor{best!15}0.061 & \cellcolor{best!40}0.045 & \cellcolor{best!40}0.063 \\
MMLPs$^\dagger$ & \cellcolor{best!15}0.063 & \cellcolor{best!00}0.066 & \cellcolor{best!00}0.084 & \cellcolor{best!00}0.071 & \cellcolor{best!00}0.062 & \cellcolor{best!00}0.051 & \cellcolor{best!00}0.066 \\
nRFGCA$^\dagger$ & \cellcolor{best!00}0.063 & \cellcolor{best!15}0.065 & \cellcolor{best!15}0.083 & \cellcolor{best!15}0.070 & \cellcolor{best!40}0.059 & \cellcolor{best!15}0.050 & \cellcolor{best!15}0.065 \\ \hline
Ours (re-init. 30) & \cellcolor{best!00}0.059 & \cellcolor{best!00}0.060 & \cellcolor{best!00}0.075 & \cellcolor{best!00}0.063 & \cellcolor{best!00}0.059 & \cellcolor{best!00}0.043 & \cellcolor{best!00}0.060 \\ 
Ours (encoder) & \cellcolor{best!00}0.055 & \cellcolor{best!00}0.056 & \cellcolor{best!00}0.068 & \cellcolor{best!00}0.060 & \cellcolor{best!00}0.057 & \cellcolor{best!00}0.040 & \cellcolor{best!00}0.056
\end{tabular}
    \label{tab:quant_percapture_test_pred}
\end{table}

\begin{table}[!h]
    \centering
    \scriptsize
    \caption{Training set image metrics on all compared methods.}
    \setlength{\tabcolsep}{3pt}
\begin{tabular}{lcccccc|c}
\textbf{PSNR$^\uparrow$} & Actor 1 & Actor 2 & Actor 3 & Actor 4 & Actor 5 & Actor 6 & Mean \\ \hline
Ours & \cellcolor{best!15}34.97 & \cellcolor{best!15}35.40 & \cellcolor{best!40}34.58 & \cellcolor{best!40}37.14 & \cellcolor{best!15}36.52 & \cellcolor{best!40}38.63 & \cellcolor{best!40}36.21 \\
MMLPs$^\dagger$ & \cellcolor{best!40}35.00 & \cellcolor{best!00}35.36 & \cellcolor{best!15}34.22 & \cellcolor{best!15}37.07 & \cellcolor{best!00}36.51 & \cellcolor{best!15}38.62 & \cellcolor{best!15}36.13 \\
nRFGCA$^\dagger$ & \cellcolor{best!00}34.69 & \cellcolor{best!40}35.58 & \cellcolor{best!00}33.42 & \cellcolor{best!00}36.84 & \cellcolor{best!40}36.53 & \cellcolor{best!00}38.56 & \cellcolor{best!00}35.94 \\
\\
\textbf{SSIM$^\uparrow$} & Actor 1 & Actor 2 & Actor 3 & Actor 4 & Actor 5 & Actor 6 & Mean \\ \hline
Ours & \cellcolor{best!40}0.939 & \cellcolor{best!15}0.956 & \cellcolor{best!40}0.949 & \cellcolor{best!40}0.956 & \cellcolor{best!00}0.934 & \cellcolor{best!15}0.969 & \cellcolor{best!40}0.950 \\
MMLPs$^\dagger$ & \cellcolor{best!00}0.939 & \cellcolor{best!00}0.956 & \cellcolor{best!15}0.947 & \cellcolor{best!15}0.955 & \cellcolor{best!15}0.934 & \cellcolor{best!00}0.969 & \cellcolor{best!40}0.950 \\
nRFGCA$^\dagger$ & \cellcolor{best!15}0.939 & \cellcolor{best!40}0.957 & \cellcolor{best!00}0.943 & \cellcolor{best!00}0.955 & \cellcolor{best!40}0.935 & \cellcolor{best!40}0.969 & \cellcolor{best!40}0.950 \\
\\
\textbf{LPIPS$^\downarrow$} & Actor 1 & Actor 2 & Actor 3 & Actor 4 & Actor 5 & Actor 6 & Mean \\ \hline
Ours & \cellcolor{best!40}0.052 & \cellcolor{best!15}0.053 & \cellcolor{best!40}0.061 & \cellcolor{best!40}0.058 & \cellcolor{best!15}0.057 & \cellcolor{best!40}0.038 & \cellcolor{best!40}0.053 \\
MMLPs$^\dagger$ & \cellcolor{best!15}0.053 & \cellcolor{best!00}0.053 & \cellcolor{best!15}0.064 & \cellcolor{best!15}0.059 & \cellcolor{best!00}0.057 & \cellcolor{best!15}0.038 & \cellcolor{best!15}0.054 \\
nRFGCA$^\dagger$ & \cellcolor{best!00}0.054 & \cellcolor{best!40}0.052 & \cellcolor{best!00}0.067 & \cellcolor{best!00}0.062 & \cellcolor{best!40}0.054 & \cellcolor{best!00}0.038 & \cellcolor{best!00}0.055
\end{tabular}
    \label{tab:quant_percapture_train}
\end{table}

\end{document}